\documentclass[10pt,twocolumn,letterpaper]{article}
\usepackage[pagenumbers]{cvpr}
\usepackage{amsmath,amssymb}
\usepackage{graphicx}
\usepackage{booktabs,tabularx}
\usepackage{multirow}
\usepackage{makecell}
\usepackage{colortbl}
\usepackage{microtype}
\usepackage{placeins}
\usepackage{cuted,caption}
\usepackage{tikz}
\usetikzlibrary{arrows.meta,positioning,fit,calc}
\newcommand{\papertitletext}{MVAgent: Per-Video Agents under Global Coordination for Multi-Video Understanding}
\newcommand{\papertitle}{MVAgent: Per-Video Agents under Global Coordination\\for Multi-Video Understanding}

\definecolor{cvprblue}{rgb}{0.21,0.49,0.74}
\definecolor{agentblue}{HTML}{355F83}
\definecolor{agentteal}{HTML}{287F78}
\definecolor{agentamber}{HTML}{946E30}
\definecolor{agentink}{HTML}{253744}
\definecolor{featuregreen}{HTML}{1B7F3B}
\definecolor{featurered}{HTML}{B42318}
\definecolor{mvagentrow}{HTML}{EDEBFF}
\newcommand{\featureyes}{\textcolor{featuregreen}{\ensuremath{\checkmark}}}
\newcommand{\featureno}{\textcolor{featurered}{\ensuremath{\times}}}
\newcommand{\resultdelta}[3]{#1\textsuperscript{\textcolor{#2}{\textbf{#3}}}}
\newcommand{\resultgains}[4]{#1\ensuremath{{}^{\text{\textcolor{featuregreen}{\textbf{#2}}}}_{\text{\textcolor{#3}{\textbf{#4}}}}}}
\newcommand{\ablbase}[1]{#1\makebox[1.4em][l]{}}
\newcommand{\abldelta}[3]{#1\makebox[1.4em][l]{\textsuperscript{\textcolor{#2}{\textbf{#3}}}}}
\newcommand{\mvagent}{MVAgent}
\newcommand{\analyze}{\textit{analyze\_videos}}
\newcommand{\watch}{\textit{watch\_videos}}
\newcommand{\observe}{\textit{observe}}
\newcommand{\finish}{\textit{finish}}
\newcommand{\answer}{\textit{answer}}

\tikzset{agentbox/.style={draw=agentblue!60,fill=agentblue!5,rounded corners=3pt,line width=.6pt,align=center,font=\sffamily\small,inner sep=6pt},agentarrow/.style={-{Latex[length=2mm]},draw=agentink!75,line width=.7pt},agentlabel/.style={font=\sffamily\scriptsize,align=center,text=agentink,fill=white,inner sep=2pt}}

\usepackage[pagebackref,breaklinks,colorlinks,allcolors=cvprblue]{hyperref}
\hypersetup{pdftitle={\papertitletext},pdfsubject={CVPR 2027 arXiv main paper}}
\makeatletter
\let\mvagentdefaultparagraph\paragraph
\renewcommand{\paragraph}{%
  \if@noskipsec \leavevmode \fi
  \par
  \@afterindenttrue
  \if@nobreak
    \everypar{}%
  \else
    \addpenalty\@secpenalty
    \vspace{1.5ex \@plus .2ex \@minus .2ex}%
  \fi
  \@ifstar
    {\@ssect{\z@}{1.5ex}{-1em}{\normalfont\normalsize\bfseries}}%
    {\@dblarg{\@sect{paragraph}{4}{\z@}{1.5ex}{-1em}%
      {\normalfont\normalsize\bfseries}}}}
\newcommand{\mvagentrestoreparagraphspacing}{\let\paragraph\mvagentdefaultparagraph}
\makeatother

\usepackage{needspace}
\def\confName{CVPR}
\def\confYear{2027}
\title{\papertitle}
\author{Kewei Wei\textsuperscript{1},
XueXiang Wen\textsuperscript{1},
Bocheng Hu\textsuperscript{1},
Weili Xu\textsuperscript{1},\\
Haiyang Yu\textsuperscript{2},
Kaiyue Zhou\textsuperscript{3}\textsuperscript{*},
Gaoang Wang\textsuperscript{1}\thanks{Corresponding authors.}\\
\textsuperscript{1}Zhejiang University, China\\\textsuperscript{2}Fudan University\\\textsuperscript{3}Chengdu Minto Tech\\
{\small\texttt{\href{mailto:12421275@zju.edu.cn}{12421275@zju.edu.cn}, \href{mailto:kyzhou@wayne.edu}{kyzhou@wayne.edu}, \href{mailto:gaoangwang@intl.zju.edu.cn}{gaoangwang@intl.zju.edu.cn}}}
}
\usepackage{tcolorbox}
\tcbuselibrary{breakable}
\usepackage{fvextra}
\fvset{fontsize=\footnotesize,breaklines=true,breakanywhere=false,breaksymbolleft={},breaksymbolright={},tabsize=2}
\definecolor{promptbar}{HTML}{D9D9D9}
\definecolor{promptfill}{HTML}{FAFAFA}
\newtcolorbox{promptbox}[3][]{breakable,colback=promptfill,colframe=black!35,
  colbacktitle=promptbar,coltitle=black,fonttitle=\bfseries\small,title={#2},
  boxrule=.5pt,arc=1pt,left=5pt,right=5pt,top=3pt,bottom=3pt,before skip=6pt,after skip=6pt,fontupper=\footnotesize,
  code={\def\promptsource{#3}},#1}
\newtcolorbox{casebox}[2]{breakable,colback=white,colframe=#1!75!black,
  colbacktitle=#1!85!black,coltitle=white,fonttitle=\bfseries\normalsize\raggedright,
  title={#2},notitle after break,before skip=6pt,after skip=6pt,boxrule=.5pt,arc=1pt,left=5pt,right=5pt,top=3pt,bottom=3pt,
  fontupper=\footnotesize}
\definecolor{memoryink}{HTML}{4F608A}
\newcommand{\casechoices}[4]{\par\smallskip
  \noindent\begin{tabularx}{\linewidth}{@{}XX@{}}
  #1 & #2\\[1pt] #3 & #4
  \end{tabularx}\par\smallskip}
\newenvironment{caseoptions}{\par\smallskip\begin{list}{}{%
  \setlength{\leftmargin}{1.6em}\setlength{\labelwidth}{1.25em}%
  \setlength{\labelsep}{.35em}\setlength{\itemsep}{1pt}%
  \setlength{\parsep}{0pt}\setlength{\topsep}{0pt}}}{\end{list}\smallskip}
\newcommand{\caseturn}[2]{\par\addvspace{3pt}\noindent
  {\color{black!20}\rule{\linewidth}{.3pt}}\par\nobreak
  \noindent{\sffamily\bfseries\fontsize{8.5}{10}\selectfont #1:}\hfill
  #2\par\nobreak}
\newcommand{\casefinal}[1]{\par\smallskip\noindent
  \fcolorbox{black!15}{black!3}{\parbox{\dimexpr\linewidth-2\fboxsep-2\fboxrule\relax}{%
    \strut\textbf{Final answer:} \texttt{#1}}}\par}
\newenvironment{failureanalysis}{\par\smallskip\begingroup\normalsize\itshape
  \noindent\textbf{Failure Analysis Note:}\ }{\par\endgroup\smallskip}
\newtcolorbox{jsonbox}{colback=black!2,colframe=black!18,boxrule=.3pt,arc=1pt,
  left=3pt,right=3pt,top=2pt,bottom=2pt,fontupper=\ttfamily\footnotesize}
\newcommand{\tightitems}{\setlength{\itemsep}{1pt}\setlength{\parsep}{0pt}%
  \setlength{\topsep}{2pt}\setlength{\partopsep}{0pt}}

\begin{document}
\maketitle
\begin{figure*}[t]
\centering
\includegraphics[width=1.0\textwidth]{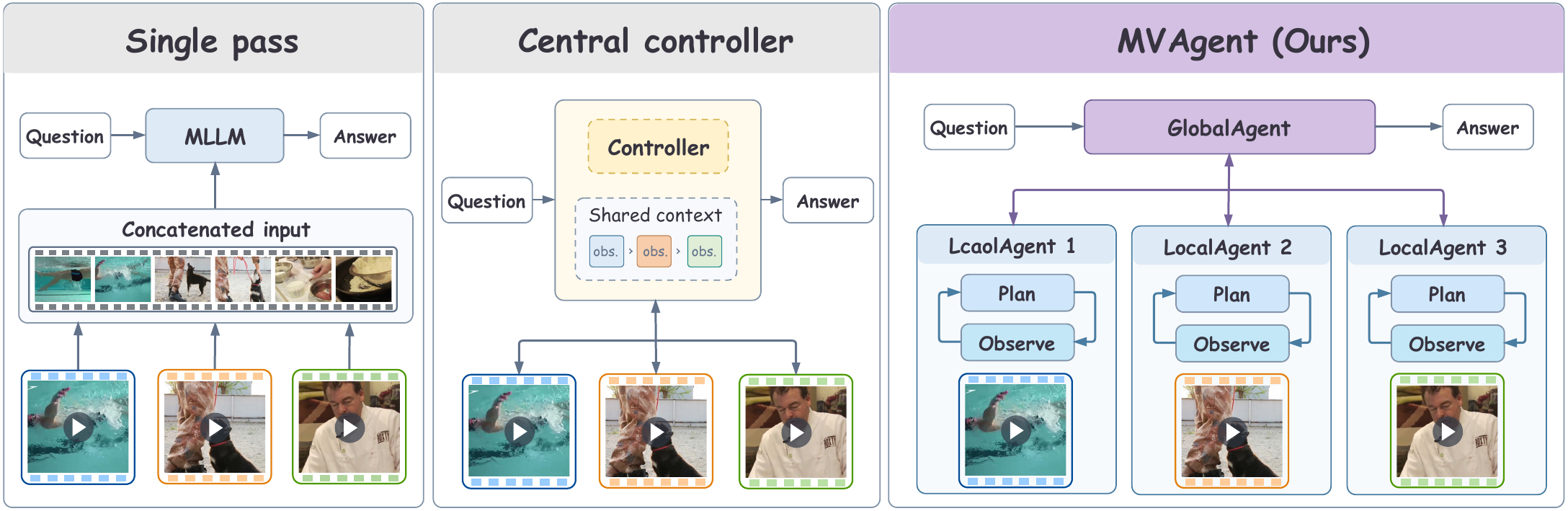}
\caption{\textbf{Multi-video understanding paradigms.} \mvagent{} assigns a dedicated LocalAgent to each video to plan its observations, while GlobalAgent coordinates cross-video reasoning.}
\label{fig:teaser}
\end{figure*}

\begin{abstract}
Multi-video understanding requires systems to combine evidence from different videos for cross-video reasoning.
Answering such questions involves two kinds of decisions.
Observation decisions must adapt to each video's content, whereas cross-video judgments require relating evidence from different videos based on the question.
Most existing agents make both kinds of decisions in one controller, whose context accumulates observations from every video.
We separate them in \textbf{\mvagent{}}, a training-free \textbf{M}ulti-\textbf{V}ideo Understanding \textbf{Agent} framework with one \mbox{LocalAgent} per video.
Each \mbox{LocalAgent} plans observations of its own video and keeps a private observation history. A \mbox{GlobalAgent} coordinates cross-video reasoning by instructing the \mbox{LocalAgents} and combining their reports with joint observation.
We evaluate \mvagent{} on CVBench, \mbox{MVU-Eval}, and CrossVid, using three Qwen3.5 models of different sizes and GPT-6-sol.
With Qwen3.5, our framework improves aggregate benchmark scores by \textbf{5.1 points} on average over Single-pass inference with the same model and by \textbf{5.0 points} over the strongest prior agent using the same model.
With GPT-6-sol, the average gain over Single-pass inference with the same model is \textbf{4.3 points}.
On the same hardware, \mvagent{} also runs \mbox{\textbf{1.8--2.7$\times$}} faster than prior agents of comparable size.
\end{abstract}

\section{Introduction}
\label{sec:intro}

Multimodal large language models (MLLMs) have advanced video understanding~\cite{an2026llavaov2}. However, much of the existing research focuses on single-video tasks~\cite{li2026crossvid,zhang2026sama}. In practice, many questions require evidence from several videos. We may compare product advertisements to choose products suited to our needs or analyze player movements and team tactics in multi-camera sports replays. Film editing also requires finding compatible clips in large collections of footage. Evidence is distributed across videos, and what to extract and how to relate it depend on both the question and the video content~\cite{peng2025mvueval,zhu2026cvbench}. Processing every video in full becomes costly as the number and duration of videos grow, although a question often depends on a few moments in each video. Which videos and moments to inspect, and in how much detail, are often difficult to determine in advance. An observation in one video can also change what must be verified in another. Answering such questions therefore requires gathering evidence step by step, with per-video analysis and cross-video judgment informing each other.

Figure~\ref{fig:teaser} compares how existing methods organize this process. Single-pass inference accommodates multiple videos through long contexts, visual compression, or question-guided frame selection, but cannot acquire further evidence when reasoning reveals a missing detail. Video-understanding agents perform multiple rounds of analysis and tool calls to compensate for information missed by Single-pass inference~\cite{wang2024videoagent,li2026lenswalk,wang2026avp}. However, existing single-video and multi-video understanding agents both rely on a single controller for planning~\cite{qiu2026agentcvr,zhao2026adseek}. When applied to multi-video understanding, the same controller must schedule observations for each video, track each video's analysis progress, and make cross-video judgments in a shared context that continually accumulates observations from all videos. However, these two kinds of decisions require different information. Cross-video judgment requires the complete question and evidence from different videos, whereas per-video observation planning mainly depends on the current goal and that video's observation history.

This difference in information needs suggests a natural division of responsibilities. Each video has a dedicated agent to plan its observations, while a global role relates evidence across videos based on the overall question. We therefore present \mvagent{}, a training-free multi-agent framework with one LocalAgent per video under global coordination. GlobalAgent uses the overall question to decide what to check in each video and combines the resulting reports. Each LocalAgent uses its observation history to plan observations and decide whether to continue or return a report. Reports and on-demand joint observation connect per-video analysis with cross-video judgment, allowing GlobalAgent to request further checks as reasoning proceeds. Intuitively, each LocalAgent's context grows only with observations of its own video, and GlobalAgent's context grows with reports rather than raw observations. All roles can share one frozen model without any role- or benchmark-specific training.

We evaluate \mvagent{} on CVBench, MVU-Eval, and CrossVid, using three Qwen3.5~\cite{qwenteam2026qwen35} models of different sizes and GPT-6-sol~\cite{openai2026gpt6sol}. Our contributions are threefold:
\begin{itemize}
\setlength{\itemsep}{3pt}
\setlength{\parsep}{0pt}
\item \textbf{Per-video agents under global coordination.} We propose \mvagent{}, a training-free framework that gives each video its own agent for observation decisions. Each LocalAgent autonomously plans how to observe its video based on GlobalAgent's instructions and the video content. It keeps a private history that later instructions can reuse. GlobalAgent maintains a global interaction history to coordinate these agents and jointly observes clips from several videos when reports miss a detail.
\item \textbf{Consistent gains and faster inference.} \mvagent{} outperforms same-model Single-pass inference in all 12 benchmark--model settings, by 5.1 points on average with Qwen3.5 and 4.3 points with \mbox{GPT-6-sol}. With Qwen3.5, it also outperforms the strongest prior agent using the same model by 5.0 points on average and runs \mbox{1.8--2.7$\times$} faster than prior agents of comparable size.
\item \textbf{Empirical validation of the architecture.} Our experiments show that \mvagent{}'s architecture enables more efficient and accurate acquisition of evidence within videos and integration across videos. LocalAgents' private histories help them select more appropriate actions.
\end{itemize}

\begin{figure*}[t]
\centering
\includegraphics[width=0.95\textwidth,pagebox=cropbox]{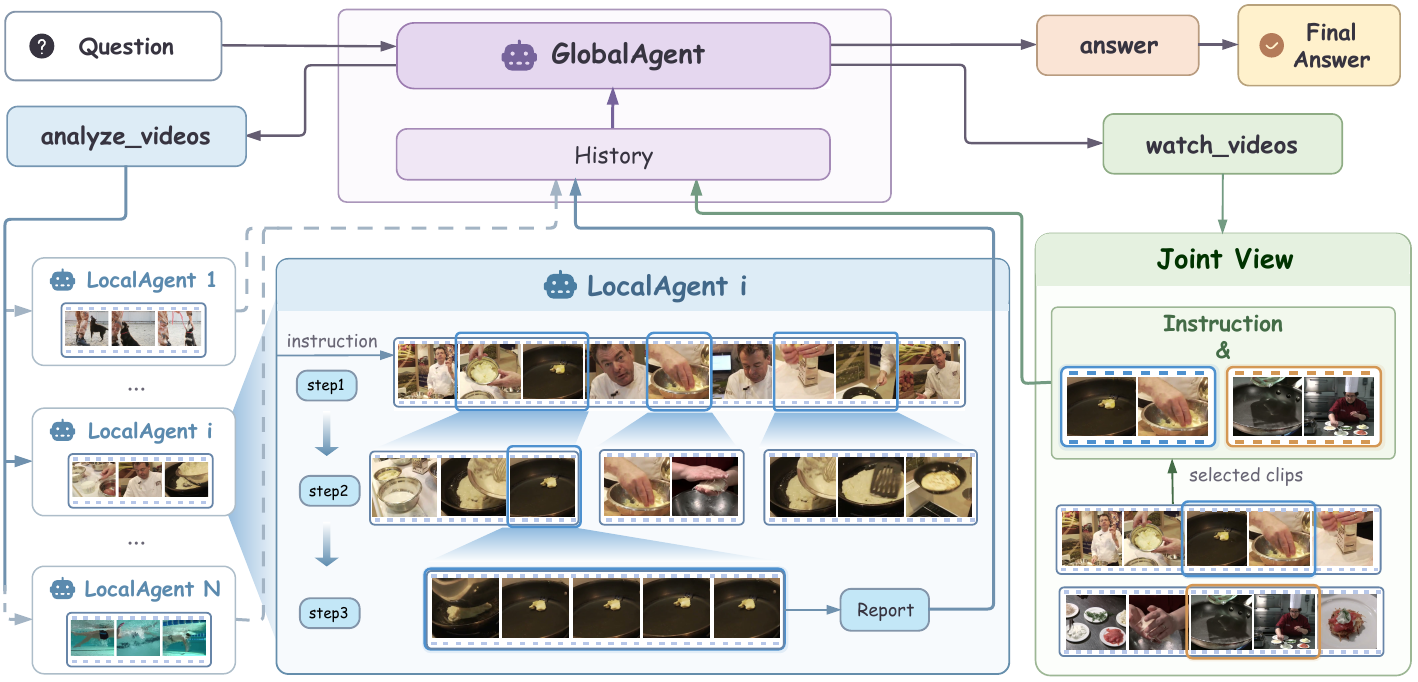}
\caption{\textbf{Overview of MVAgent.} GlobalAgent coordinates one LocalAgent per video and supports joint observation of selected clips across videos. Local reports and joint observations update global history to guide further observation and cross-video reasoning, while detailed local observations remain private.}
\label{fig:pipeline}
\end{figure*}

\section{Related Work}
\label{sec:related}

\begin{table}[t]
\centering
\caption{\textbf{Video agent comparison.}
Two-level: a global agent with per-video agents. Joint obs.: one visual call covers several videos.
\textsuperscript{*} Full inference pipeline unavailable.}
\label{tab:systems}
\small
\setlength{\tabcolsep}{1pt}
\renewcommand{\arraystretch}{1.14}
\begin{tabularx}{\columnwidth}{>{\raggedright\arraybackslash}p{64pt}*{4}{>{\centering\arraybackslash}X}}
\toprule
\textbf{Method} & \textbf{Planning} & \textbf{History} & \textbf{\makecell{Joint\\obs.}} & \textbf{\makecell{Training-\\free}} \\
\midrule
\multicolumn{5}{l}{\textit{Single-video agents}}\\[-0.2ex]
AVP~\cite{wang2026avp} & Central & Central & \textendash & \featureyes \\
LensWalk~\cite{li2026lenswalk} & Central & Central & \textendash & \featureyes \\
Video-RSI~\cite{luo2026videorsi} & Central & Central & \textendash & \featureyes \\
\midrule
\multicolumn{5}{l}{\textit{Multi-video agents}}\\[-0.2ex]
MAGNET\textsuperscript{*}~\cite{chowdhury2025magnet} & Fixed & None & \featureno & \featureno \\
SAMA\textsuperscript{*}~\cite{zhang2026sama} & Central & Central & \featureno & \featureyes \\
AgentCVR~\cite{qiu2026agentcvr} & Central & Central & \featureyes & \featureno \\
AdSeek\textsuperscript{*}~\cite{zhao2026adseek} & Central & Central & \featureno & \featureno \\
\midrule
\rowcolor{black!10}
\textbf{\mvagent{}} & \textbf{Two-level} & \textbf{Two-level} & \featureyes & \featureyes \\
\bottomrule
\end{tabularx}
\end{table}

\paragraph{Context organization in agent systems.}
Agents following the ReAct paradigm~\cite{yao2023react} interleave reasoning and actions over multiple rounds, and each round adds new observations to their contexts. Long inputs and accumulated observations require agents to organize their working contexts within finite budgets~\cite{zhao2024longagent,yin2026videoarm}. The challenge is to keep these contexts compact without losing details that later reasoning may need~\cite{ma2025drvideo,li2026hmars}. Text agents move information between the context window and external storage~\cite{packer2023memgpt} or keep compressed memories of long documents~\cite{lee2024readagent}. Multi-agent systems distribute document chunks or bounded memory regions among local roles. Higher-level roles coordinate access to these records and integrate the resulting evidence~\cite{zhang2024chainofagents}. In video understanding, question-aware frame selection~\cite{buch2025ffs} and query-adaptive hierarchical representations~\cite{wang2025videotree} construct visual context around a question. Retrieval-based approaches organize videos as documents that can be iteratively augmented, or as multi-granular databases for agentic search~\cite{zhang2025dvd}. Short- and long-term memory compress visual representations~\cite{song2024moviechat}. Agentic methods also construct hierarchical multimodal memory during execution. \mvagent{} brings this division among local roles to video, with one LocalAgent per video. Unlike a document chunk, a video often cannot be viewed in full at the sampling rate that brief details require, and each LocalAgent therefore also decides how to observe its video.

\paragraph{Single-video understanding.}
Single-video benchmarks assess temporal, long-context, and general video question answering~\cite{li2024mvbench,fu2025videomme,zhou2025mlvu,wu2024longvideobench}. A question may hinge on a brief visual detail or require relating events separated in time, motivating observation strategies that balance temporal coverage with detailed inspection~\cite{wang2026avp,li2026lenswalk}. Agentic approaches incorporate iterative retrieval and segment exploration into reasoning, acquiring further evidence based on intermediate observations~\cite{wang2024videoagent,yang2025vca}. Training-free methods adapt visual queries to the details to be checked, with LensWalk controlling temporal scope and sampling rate. Observation policies can also be learned through reinforcement learning~\cite{zhang2026eva}. Video-RSI~\cite{luo2026videorsi} uses execution feedback to evolve the executable reasoning harness. Multi-agent approaches combine modality-specific analysis or multi-round team discussion~\cite{kugo2025videomultiagents,chen2025lvagent}. Other systems separate planning, grounding, perception, and reflection into specialized roles that exchange results during reasoning~\cite{yan2026symphony}. \mvagent{} runs one LocalAgent per video and divides work by video rather than by modality or role.

\paragraph{Multi-video understanding.}
Cross-video questions require associating events, states, and identities across videos, where differences in viewpoints and local timelines complicate evidence alignment~\cite{peng2025mvueval,li2026crossvid}. Benchmarks examine differences and state changes~\cite{nagarajan2024stepdiff,qiu2025videosetdiff,wei2026m3verse}, multi-view relations~\cite{yeh2026allangles}, audio-visual retrieval and temporal grounding~\cite{chowdhury2025magnet}, and broader capabilities~\cite{bai2026mvpbench}. CVBench~\cite{zhu2026cvbench}, MVU-Eval, and CrossVid assess complementary forms of cross-video perception and reasoning. Methods explore video selection, temporal localization, and evidence aggregation as complementary parts of multi-video reasoning. VidDiff~\cite{burgess2025viddiff} and MAGNET use staged workflows for fine-grained action differencing and cross-video audio-visual question answering, respectively. MAGNET localizes evidence and produces partial responses through video-specific audio-visual roles before centralized aggregation. VideoForest~\cite{meng2025videoforest} focuses on person-centric cross-video understanding. SAMA~\cite{zhang2026sama} uses frozen models with a central planner that manages video-indexed evidence, issues tool queries, and rechecks conflicting evidence. AgentCVR~\cite{qiu2026agentcvr} trains a central controller through script-simulated reinforcement learning to select videos and set observation intervals and queries. Modality-specific agents return evidence to this controller. AdSeek~\cite{zhao2026adseek} learns multi-round acquisition for e-commerce videos. Our focus is globally coordinated per-video analysis. Unlike MAGNET, whose per-video models run a single fixed pass, each LocalAgent can plan its own observations over several steps and keeps its history across requests.

\section{MVAgent}
\label{sec:framework}

We formalize multi-video understanding as
\begin{equation}
\mathcal{M}\bigl((V_1,\ldots,V_N),\,Q\bigr)\rightarrow O,\quad N\geq 2,
\label{eq:multi-video-task}
\end{equation}
where $\mathcal{M}$ denotes the system, $(V_1,\ldots,V_N)$ is the ordered input collection of $N$ videos, $Q$ is the question, and $O$ is the output, whose format follows the task. Input videos may be related in various ways~\cite{zhu2026cvbench,peng2025mvueval,li2026crossvid}. The framework must acquire and combine evidence to answer the current question, retaining the video identity and local timestamps of each observation.

\Needspace{1.5\baselineskip}
We separate cross-video judgment from per-video analysis to give each LocalAgent control over its observations. As shown in Figure~\ref{fig:pipeline}, for a question $Q$ and its input videos, we create one GlobalAgent and one LocalAgent per video. Each agent maintains its own decision context and history. GlobalAgent issues question-specific instructions to selected LocalAgents, which plan their observations and return reports. GlobalAgent can also select clips for joint observation. It combines reports and observations to decide whether to gather more evidence or answer. Both GlobalAgent and LocalAgent consist of a text-only planning model and a visual model. The planning model selects actions, writes instructions, and composes the final answer, while the visual model performs visual observation. All agents share one planning model and one visual model, which can be the same frozen model; Table~\ref{tab:model-allocation} compares different allocations.


\begin{table*}[t]
\centering
\caption{\textbf{Multi-video understanding results.} Single-pass uses one model call. \textbf{Bold}/\underline{underline} mark the best/second-best results within each model group; for Time, bold and underline compare agent methods only. Score gains are in percentage points (pp): vs.\ Single-pass above, best same-model agent below. Time: mean s/question; MVAgent superscripts report speedup over the fastest same-model agent.}
\label{tab:main-detail}
\footnotesize
\setlength{\tabcolsep}{1.1pt}
\renewcommand{\arraystretch}{1.15}
\begin{tabularx}{\textwidth}{>{\raggedright\arraybackslash}p{68pt}!{\tikz[remember picture,overlay]{\coordinate (table-model-v1);}}*{3}{>{\centering\arraybackslash}X}!{\tikz[remember picture,overlay]{\coordinate (table-model-v4);}}c!{\tikz[remember picture,overlay]{\coordinate (table-model-v5);}}*{8}{>{\centering\arraybackslash}X}!{\tikz[remember picture,overlay]{\coordinate (table-model-v13);}}c!{\tikz[remember picture,overlay]{\coordinate (table-model-v14);}}*{4}{>{\centering\arraybackslash}X}!{\tikz[remember picture,overlay]{\coordinate (table-model-v18);}}c!{\tikz[remember picture,overlay]{\coordinate (table-model-v19);}}c}
\toprule
\noalign{\hbox{\tikz[remember picture,overlay]{\coordinate (table-model-top);}}}
\multirow{2}{=}[-0.7ex]{\centering\textbf{Method}} & \multicolumn{4}{c}{$\uparrow$~\textbf{CVBench}~\cite{zhu2026cvbench}} & \multicolumn{9}{c}{$\uparrow$~\textbf{MVU-Eval}~\cite{peng2025mvueval}} & \multicolumn{5}{c}{$\uparrow$~\textbf{CrossVid}~\cite{li2026crossvid}} & \multirow{2}{*}[-0.7ex]{\textbf{Time (s)}~$\downarrow$}\\
\cmidrule{2-19}
\noalign{\hbox{\tikz[remember picture,overlay]{\coordinate (table-model-subtop);}}}
 & \textbf{Obj} & \textbf{Evt} & \textbf{Reas} & \textbf{All} & \textbf{OR} & \textbf{SU} & \textbf{Cnt} & \textbf{Cmp} & \textbf{KIR} & \textbf{ICL} & \textbf{RAG} & \textbf{TR} & \textbf{All} & \textbf{Cmp} & \textbf{Tmp} & \textbf{View} & \textbf{Free} & \textbf{All} & \\
\noalign{\hbox{\tikz[remember picture,overlay]{\coordinate (table-model-header-bottom);}}}
\midrule
\rowcolor{black!8}
\multicolumn{20}{l}{\textit{AgentCVR-8B}~\cite{qiu2026agentcvr}}\\
\noalign{\hbox{\tikz[remember picture,overlay]{\coordinate (table-model-g1-top);}}}
\hspace{4pt}AgentCVR & 56.5 & 58.1 & 55.9 & 56.9 & 41.3 & 46.4 & 54.6 & 48.1 & 36.3 & 28.0 & 40.4 & 52.0 & 44.0 & 48.4 & 22.1 & 36.8 & 54.8 & 38.8 & 153.8\\
\noalign{\hbox{\tikz[remember picture,overlay]{\coordinate (table-model-g1-bottom);}}}
\midrule
\rowcolor{black!8}
\multicolumn{20}{l}{\textit{Qwen3.5-9B}}\\
\noalign{\hbox{\tikz[remember picture,overlay]{\coordinate (table-model-g2-top);}}}
\hspace{4pt}\textcolor{black!55}{\textit{Single-pass}} & \textcolor{black!55}{\underline{60.9}} & \textcolor{black!55}{\underline{62.7}} & \textcolor{black!55}{\textbf{62.3}} & \textcolor{black!55}{\underline{62.1}} & \textcolor{black!55}{38.1} & \textcolor{black!55}{\underline{47.5}} & \textcolor{black!55}{49.3} & \textcolor{black!55}{\underline{66.7}} & \textcolor{black!55}{\underline{40.2}} & \textcolor{black!55}{\underline{25.0}} & \textcolor{black!55}{44.8} & \textcolor{black!55}{\underline{54.4}} & \textcolor{black!55}{46.3} & \textcolor{black!55}{30.0} & \textcolor{black!55}{\underline{28.0}} & \textcolor{black!55}{31.5} & \textcolor{black!55}{\textbf{50.1}} & \textcolor{black!55}{\underline{31.7}} & \textcolor{black!55}{9.5}\\
\cmidrule(lr){1-20}
\hspace{4pt}LensWalk~\cite{li2026lenswalk} & 55.2 & 58.1 & 55.1 & 56.3 & 34.9 & 35.2 & 34.8 & 61.5 & 19.9 & 9.1 & 41.9 & 48.5 & 36.3 & 29.8 & 25.8 & 13.9 & \underline{39.9} & 26.4 & 158.7\\
\hspace{4pt}Video-RSI~\cite{luo2026videorsi} & 57.7 & 59.7 & \underline{56.5} & 58.0 & \underline{42.1} & 42.5 & \underline{53.3} & \textbf{70.4} & \textbf{45.2} & 24.4 & \textbf{46.9} & 48.3 & \underline{46.7} & \underline{34.4} & 25.0 & \underline{32.4} & 30.3 & 30.8 & \underline{116.4}\\
\rowcolor{mvagentrow}
\rule[-0.7ex]{0pt}{3ex}\hspace{4pt}\textbf{MVAgent} & \textbf{64.9} & \textbf{66.5} & \textbf{62.3} & \resultgains{\textbf{64.5}}{+2.4}{featuregreen}{+6.5} & \textbf{47.6} & \textbf{57.0} & \textbf{67.0} & 62.2 & 34.9 & \textbf{28.0} & \underline{45.4} & \textbf{60.6} & \resultgains{\textbf{50.5}}{+4.3}{featuregreen}{+3.9} & \textbf{47.0} & \textbf{33.6} & \textbf{33.9} & 28.6 & \resultgains{\textbf{38.5}}{+6.8}{featuregreen}{+7.8} & \resultdelta{\textbf{59.6}}{featuregreen}{1.95$\times$}\\
\noalign{\hbox{\tikz[remember picture,overlay]{\coordinate (table-model-g2-bottom);}}}
\midrule
\rowcolor{black!8}
\multicolumn{20}{l}{\textit{Qwen3.5-35B-A3B}}\\
\noalign{\hbox{\tikz[remember picture,overlay]{\coordinate (table-model-g3-top);}}}
\hspace{4pt}\textcolor{black!55}{\textit{Single-pass}} & \textcolor{black!55}{62.9} & \textcolor{black!55}{63.8} & \textcolor{black!55}{\textbf{65.4}} & \textcolor{black!55}{\underline{64.2}} & \textcolor{black!55}{41.3} & \textcolor{black!55}{\underline{51.4}} & \textcolor{black!55}{\underline{59.5}} & \textcolor{black!55}{\textbf{78.5}} & \textcolor{black!55}{\underline{43.1}} & \textcolor{black!55}{22.0} & \textcolor{black!55}{46.3} & \textcolor{black!55}{53.1} & \textcolor{black!55}{49.2} & \textcolor{black!55}{\underline{39.9}} & \textcolor{black!55}{28.5} & \textcolor{black!55}{19.9} & \textcolor{black!55}{\textbf{47.6}} & \textcolor{black!55}{33.2} & \textcolor{black!55}{11.3}\\
\cmidrule(lr){1-20}
\hspace{4pt}LensWalk & 53.6 & 58.1 & 53.8 & 55.4 & 29.4 & 31.8 & 32.2 & 58.5 & 11.4 & 14.0 & 35.7 & 51.7 & 33.7 & 23.4 & 25.2 & 10.0 & 33.2 & 22.2 & \underline{217.3}\\
\hspace{4pt}Video-RSI & \underline{64.5} & \underline{64.1} & 59.4 & 62.4 & \underline{46.0} & 49.2 & 52.9 & \underline{75.6} & \textbf{45.6} & \underline{23.8} & \textbf{51.9} & \underline{60.9} & \underline{51.4} & 38.8 & \underline{32.1} & \underline{33.4} & 27.3 & \underline{34.6} & 307.2\\
\rowcolor{mvagentrow}
\rule[-0.7ex]{0pt}{3ex}\hspace{4pt}\textbf{MVAgent} & \textbf{69.0} & \textbf{70.3} & \underline{63.9} & \resultgains{\textbf{67.5}}{+3.3}{featuregreen}{+5.1} & \textbf{46.8} & \textbf{54.2} & \textbf{65.2} & 67.4 & 39.1 & \textbf{28.0} & \underline{48.7} & \textbf{73.5} & \resultgains{\textbf{54.3}}{+5.1}{featuregreen}{+2.9} & \textbf{46.0} & \textbf{40.5} & \textbf{38.3} & \underline{44.5} & \resultgains{\textbf{42.7}}{+9.4}{featuregreen}{+8.1} & \resultdelta{\textbf{122.4}}{featuregreen}{1.78$\times$}\\
\noalign{\hbox{\tikz[remember picture,overlay]{\coordinate (table-model-g3-bottom);}}}
\midrule
\rowcolor{black!8}
\multicolumn{20}{l}{\textit{Qwen3.5-27B}}\\
\noalign{\hbox{\tikz[remember picture,overlay]{\coordinate (table-model-g4-top);}}}
\hspace{4pt}\textcolor{black!55}{\textit{Single-pass}} & \textcolor{black!55}{\underline{68.5}} & \textcolor{black!55}{71.1} & \textcolor{black!55}{64.9} & \textcolor{black!55}{68.1} & \textcolor{black!55}{\textbf{51.6}} & \textcolor{black!55}{54.7} & \textcolor{black!55}{57.7} & \textcolor{black!55}{\textbf{83.0}} & \textcolor{black!55}{\textbf{48.0}} & \textcolor{black!55}{\underline{28.0}} & \textcolor{black!55}{49.3} & \textcolor{black!55}{61.1} & \textcolor{black!55}{53.8} & \textcolor{black!55}{\textbf{51.3}} & \textcolor{black!55}{31.1} & \textcolor{black!55}{25.6} & \textcolor{black!55}{\textbf{66.6}} & \textcolor{black!55}{41.7} & \textcolor{black!55}{17.9}\\
\cmidrule(lr){1-20}
\hspace{4pt}LensWalk & 61.3 & 61.6 & 63.1 & 62.1 & 32.5 & 40.2 & 25.6 & 72.6 & 16.4 & 6.7 & 41.0 & 54.4 & 36.6 & 36.2 & 36.9 & 17.6 & \underline{52.2} & 34.3 & \underline{249.5}\\
\hspace{4pt}Video-RSI & 67.7 & \underline{71.4} & \underline{65.4} & \underline{68.2} & 50.0 & \underline{57.0} & \underline{67.0} & \underline{76.3} & 43.8 & 18.9 & \underline{50.1} & \underline{68.4} & \underline{54.8} & \underline{50.1} & \underline{45.9} & \underline{31.4} & 44.4 & \underline{44.5} & 266.3\\
\rowcolor{mvagentrow}
\rule[-0.7ex]{0pt}{3ex}\hspace{4pt}\textbf{MVAgent} & \textbf{69.4} & \textbf{71.6} & \textbf{70.2} & \resultgains{\textbf{70.5}}{+2.4}{featuregreen}{+2.3} & \underline{50.8} & \textbf{58.7} & \textbf{67.0} & 73.3 & \underline{45.9} & \textbf{28.7} & \textbf{51.0} & \textbf{78.3} & \resultgains{\textbf{58.2}}{+4.3}{featuregreen}{+3.4} & 50.0 & \textbf{53.1} & \textbf{42.4} & 48.4 & \resultgains{\textbf{49.3}}{+7.6}{featuregreen}{+4.8} & \resultdelta{\textbf{121.8}}{featuregreen}{2.05$\times$}\\
\noalign{\hbox{\tikz[remember picture,overlay]{\coordinate (table-model-g4-bottom);}}}
\midrule
\rowcolor{black!8}
\multicolumn{20}{l}{\textit{GPT-6-sol}}\\
\noalign{\hbox{\tikz[remember picture,overlay]{\coordinate (table-model-g0-top);}}}
\hspace{4pt}\textcolor{black!55}{\textit{Single-pass}} & \textcolor{black!55}{\underline{75.4}} & \textcolor{black!55}{\underline{76.8}} & \textcolor{black!55}{\underline{74.4}} & \textcolor{black!55}{\underline{75.5}} & \textcolor{black!55}{\underline{48.4}} & \textcolor{black!55}{\underline{59.8}} & \textcolor{black!55}{\underline{83.3}} & \textcolor{black!55}{\underline{80.0}} & \textcolor{black!55}{\underline{51.6}} & \textcolor{black!55}{\textbf{62.8}} & \textcolor{black!55}{\textbf{56.3}} & \textcolor{black!55}{\underline{88.5}} & \textcolor{black!55}{\underline{67.7}} & \textcolor{black!55}{\underline{65.1}} & \textcolor{black!55}{\underline{53.1}} & \textcolor{black!55}{\underline{43.8}} & \textcolor{black!55}{\textbf{65.4}} & \textcolor{black!55}{\underline{57.3}} & \textcolor{black!55}{--}\\
\cmidrule(lr){1-20}
\rowcolor{mvagentrow}
\rule[-0.7ex]{0pt}{3ex}\hspace{4pt}\textbf{MVAgent} & \textbf{82.7} & \textbf{77.8} & \textbf{75.7} & \resultdelta{\textbf{78.2}}{featuregreen}{+2.7} & \textbf{57.9} & \textbf{62.0} & \textbf{88.1} & \textbf{81.5} & \textbf{64.1} & \underline{60.4} & \textbf{56.3} & \textbf{89.5} & \resultdelta{\textbf{71.2}}{featuregreen}{+3.5} & \textbf{67.6} & \textbf{68.2} & \textbf{51.6} & \underline{60.8} & \resultdelta{\textbf{63.9}}{featuregreen}{+6.6} & \textcolor{black!55}{--}\\
\noalign{\hbox{\tikz[remember picture,overlay]{\coordinate (table-model-g0-bottom);}}}
\noalign{\hbox{\tikz[remember picture,overlay]{\coordinate (table-model-bottom);}}}
\bottomrule
\end{tabularx}
\tikz[remember picture,overlay]{
\foreach \x in {1,5,14,19} {
\draw[black!85,line width=0.65pt] (table-model-v\x |- table-model-top) -- (table-model-v\x |- table-model-header-bottom);
}
\foreach \x in {4,13,18} {
\draw[black!85,line width=0.65pt] (table-model-v\x |- table-model-subtop) -- (table-model-v\x |- table-model-header-bottom);
}
\foreach \x in {1,4,5,13,14,18,19} {
\draw[black!85,line width=0.65pt] (table-model-v\x |- table-model-g0-top) -- (table-model-v\x |- table-model-g0-bottom);
}
\foreach \x in {1,4,5,13,14,18,19} {
\draw[black!85,line width=0.65pt] (table-model-v\x |- table-model-g1-top) -- (table-model-v\x |- table-model-g1-bottom);
}
\foreach \x in {1,4,5,13,14,18,19} {
\draw[black!85,line width=0.65pt] (table-model-v\x |- table-model-g2-top) -- (table-model-v\x |- table-model-g2-bottom);
}
\foreach \x in {1,4,5,13,14,18,19} {
\draw[black!85,line width=0.65pt] (table-model-v\x |- table-model-g3-top) -- (table-model-v\x |- table-model-g3-bottom);
}
\foreach \x in {1,4,5,13,14,18,19} {
\draw[black!85,line width=0.65pt] (table-model-v\x |- table-model-g4-top) -- (table-model-v\x |- table-model-g4-bottom);
}
\foreach \x in {1,5,14,19} {
\draw[black,line width=1.2pt] (table-model-v\x |- table-model-top) -- (table-model-v\x |- table-model-header-bottom);
\foreach \g in {0,1,2,3,4} {
\draw[black,line width=1.2pt] (table-model-v\x |- table-model-g\g-top) -- (table-model-v\x |- table-model-g\g-bottom);
}
}
}
\end{table*}

\subsection{GlobalAgent}
For each question $Q$, we assign one GlobalAgent to coordinate the overall analysis based on the question and accumulated evidence.

\paragraph{Actions.} GlobalAgent selects an action and parameters as
\begin{equation}
(u,\theta)=\pi_{\mathrm{G}}(Q,\mathcal{S},H_{\mathrm{G}}^k),
\label{eq:global-decision}
\end{equation}
where $\pi_{\mathrm{G}}$ denotes the global decision rule, $u$ and $\theta$ specify the action and its parameters, $\mathcal{S}$ lists video metadata, including identifiers and durations, and $H_{\mathrm{G}}^k$ is the global history. Three actions are available:

\begin{itemize}
\setlength{\itemsep}{3pt}
\setlength{\parsep}{0pt}
\item \mbox{\textbf{analyze\_videos.}} \mbox{GlobalAgent} selects one or more videos and sends each video's \mbox{LocalAgent} an instruction $I$, specifying what to check or compare. The instruction must give the LocalAgent enough information to complete the task using only its video, without reading the complete question or global history. A request is one call to a LocalAgent with an instruction. The selected \mbox{LocalAgents} can analyze their videos in parallel, each planning its own observations and returning a report.

\item \mbox{\textbf{watch\_videos.}} Reports returned by \mbox{\analyze{}} may omit fine-grained visual details needed, for example, to determine whether the people wearing red in two videos are the same person. For joint observation, \mbox{GlobalAgent}~selects at least two distinct videos and one time interval within each video's duration, then provides a \mbox{comparison} instruction that names the selected videos and specifies the relationship to examine. The visual model inspects the selected clips together in the specified video order and returns an observation.

\item \textbf{answer.} When GlobalAgent judges the evidence sufficient to answer the question or reaches its round limit, it synthesizes a final answer in the required format.
\end{itemize}

\paragraph{Global history.}
Global history retains the acquisition decisions and returned results for the current question to support subsequent video selection, instruction generation, and cross-video judgment. Each acquisition action and its returned result $R$ produce the next global history version:
\begin{equation}
H_{\mathrm{G}}^0=\varnothing,\quad H_{\mathrm{G}}^{k+1}=H_{\mathrm{G}}^k\oplus(u,\theta,R).
\label{eq:global-history}
\end{equation}
Here $k$ indexes global acquisition rounds, $R$ contains LocalAgent reports or visual observations together with execution feedback, and $\oplus$ appends a record in execution order. GlobalAgent uses this evidence to form later instructions, which LocalAgents execute using retained local histories.

\subsection{LocalAgent}
Each input video $V_i$ is assigned a LocalAgent that remains responsible for it throughout the question. Guided by GlobalAgent's instruction $I$ and its private local history $H_i^t$, it autonomously decides how to observe the video and whether further observation is needed. It receives the instruction rather than the complete question or global history. It exchanges instructions and reports only with GlobalAgent, without communicating with other LocalAgents.

\paragraph{Actions.} With video metadata fixed, LocalAgent selects an action and its parameters from the current instruction $I$ and local history $H_i^t$:
\begin{equation}
(u,\theta)=\pi_{\mathrm{L}}(I,H_i^t),
\label{eq:video-decision}
\end{equation}
where $\pi_{\mathrm{L}}$ is the local decision rule of a frozen model.
\begin{itemize}
\setlength{\itemsep}{3pt}
\setlength{\parsep}{0pt}
\item \textbf{observe.} Inspired by AVP~\cite{wang2026avp}, LocalAgent selects a visual query $q$, one or more video time intervals $\tau$, and a sampling rate $f$ (fps) based on what needs verification. These form the action parameters $\theta=(q,\tau,f)$. For example, a long video can be inspected in segments, or a brief action revisited with denser sampling. The visual model inspects each interval separately, receiving only the current visual query, selected clip, and video timestamps. The returned observations, including any uncertainty, guide the next decision.

\item \textbf{finish.} When LocalAgent judges the evidence sufficient for the current instruction or reaches its observation limit, it summarizes relevant evidence and unresolved uncertainty in a textual report that GlobalAgent can understand without reading local observation records.
\end{itemize}

\paragraph{Local history.}
Local history provides a basis for subsequent local action selection. Prior instructions, observation actions, sampling parameters, observations, and execution feedback help the LocalAgent identify what it has already checked and what still requires investigation, and are therefore retained. \textit{Finish} reports summarize existing evidence and add no new observations, so they are excluded from local history. For video $V_i$, $H_i^t$ denotes the current local history, with $t$ counting its observation steps across requests within the question. After an \observe{} call, its instruction, action, parameters, and feedback $o$ are recorded:
\begin{equation}
H_i^0=\varnothing,\quad H_i^{t+1}=H_i^t\oplus(I,u,\theta,o).
\label{eq:video-history}
\end{equation}
Here $o$ contains returned observations and execution feedback. Records are grouped by request and persist when a new instruction arrives, allowing later requests to reuse earlier observations without necessarily observing again. The observations behind earlier reports remain available when a later instruction asks for a different detail.

\section{Experiments}
\label{sec:experiments}

\subsection{Experimental Setup}
\label{sec:experimental-setup}
We evaluate \mvagent{} on three comprehensive multi-video benchmarks: CVBench~\cite{zhu2026cvbench}, MVU-Eval~\cite{peng2025mvueval}, and CrossVid~\cite{li2026crossvid}. Together, these benchmarks span diverse tasks and question formats, including object and event understanding, temporal and cross-view relations, and open-ended synthesis. To improve evaluation efficiency, we use a stratified random sample of 2,500 CrossVid questions that approximately preserves its task distribution. We follow each benchmark's official scoring protocol. For open-ended questions, we use DeepSeek-V4.1-Flash~\cite{deepseek2026flashdocs} as the judge and apply the official scoring script and prompt.

We use public implementations for the agent baselines.
\begin{itemize}
\setlength{\itemsep}{3pt}
\setlength{\parsep}{0pt}
\item \textbf{Single-pass evaluation.} We evaluate the open-source Qwen3.5-9B~\cite{qwenteam2026qwen35}, Qwen3.5-35B-A3B, and Qwen3.5-27B, alongside GPT-6-sol~\cite{openai2026gpt6sol}, using Single-pass inference. We use the official benchmark question and answer instructions. We allow at most 512 frames in total per question and a maximum frame side length of 448 pixels. Thinking is disabled for Qwen3.5 and set to \texttt{low} for GPT-6-sol.

\item \textbf{Single-video agents.} LensWalk~\cite{li2026lenswalk} and Video-RSI~\cite{luo2026videorsi} use the same local Qwen3.5 models internally, with thinking disabled. Following AgentCVR~\cite{qiu2026agentcvr}, we adapt the single-video input interfaces by concatenating videos in benchmark order and providing each video's ID and temporal boundaries. For Video-RSI, we evaluate only the final evolved version released by its authors. We lightly adapt its answer prompts and validation code to support benchmark answer formats beyond its original multiple-choice interface. We otherwise follow the official settings.

\item \textbf{Multi-video agents.} We evaluate the released trained AgentCVR-8B with its official framework and settings. For \mvagent{}'s three local Qwen3.5 configurations, all roles use the same frozen model with thinking disabled. For its GPT-6-sol configuration, we set thinking to \texttt{low}. The system chooses sampling within a 512-frame limit per visual call, without requiring the full budget. Joint observation covers up to four distinct videos. GlobalAgent runs at most five rounds per question, and each LocalAgent takes at most five steps per request.
\end{itemize}

\subsection{Main Results}
\paragraph{Overall performance.}
Table~\ref{tab:main-detail} shows that \mvagent{} outperforms the same-model Single-pass baseline in all 12 benchmark--model settings using three Qwen3.5 models of different sizes and GPT-6-sol. The gains are 2.4--9.4 points across the nine Qwen3.5 benchmark--model settings and 2.7--6.6 points across the three GPT-6-sol benchmark settings, spanning a 9B open-weight model through \mbox{GPT-6-sol}. This shows that \mvagent{} is effective across model sizes. In all nine Qwen3.5 settings, \mvagent{} exceeds the strongest same-model agent baseline by 2.3--8.1 points. Training-free \mvagent{} with Qwen3.5-9B (\mbox{\mvagent{}-9B}) also surpasses the larger Single-pass Qwen3.5-35B-A3B on all three benchmarks and the reinforcement-trained AgentCVR-8B~\cite{qiu2026agentcvr} on CVBench~\cite{zhu2026cvbench} and MVU-Eval~\cite{peng2025mvueval}. It remains competitive on CrossVid~\cite{li2026crossvid}, where AgentCVR received targeted reinforcement learning.

\begin{figure}[!t]
\begin{minipage}[t]{\columnwidth}
\vspace{0pt}
\centering
\captionof{table}{\textbf{Online inference statistics.} Mean calls and frames per question; mean output tokens per call. Results pool three Qwen3.5 model sizes, except AgentCVR-8B~\cite{qiu2026agentcvr}.}
\label{tab:execution-stats}
\small
\setlength{\tabcolsep}{4pt}
\renewcommand{\arraystretch}{1}
\begin{tabular*}{\columnwidth}{@{\extracolsep{\fill}}lcccc@{}}
\toprule
\multirow[c]{2}{*}{Method} & \multicolumn{2}{c}{Calls} & \multicolumn{1}{c}{\multirow[c]{2}{*}{Frames}} & \multicolumn{1}{c}{\multirow[c]{2}{*}{Out. tok.}}\\
\cmidrule(lr){2-3}
 & \multicolumn{1}{c}{Plan} & \multicolumn{1}{c}{Visual} & & \\
\midrule
LensWalk~\cite{li2026lenswalk} & 4.6 & 14.1 & 151.9 & 234.5\\
Video-RSI~\cite{luo2026videorsi} & 3.9 & 2.1 & 41.9 & 842.1\\
AgentCVR-8B & 5.3 & 1.7 & 71.9 & 1,005.4\\
\midrule
\textbf{\mvagent{}} & 9.0 & 3.6 & 109.9 & 137.5\\
\bottomrule
\end{tabular*}

\end{minipage}
\par\medskip
\begin{minipage}[t]{\columnwidth}
\vspace{0pt}
\centering
\includegraphics[width=\linewidth]{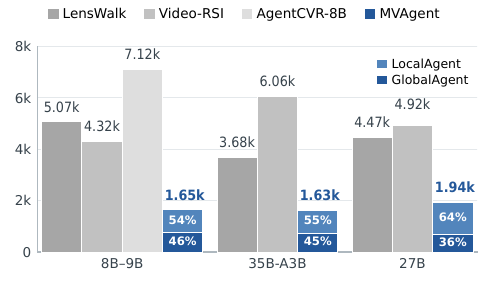}
\caption{\textbf{Mean output tokens per question.} \mvagent{} bars separate GlobalAgent and LocalAgent generation.}
\label{fig:generation-cost}
\end{minipage}
\end{figure}

\paragraph{Execution efficiency.}
\looseness=-1 On the same hardware, \mvagent{} retains its performance advantage with 1.8--2.7$\times$ speedups over agent baselines of comparable size. This speedup comes mainly from the token efficiency of \mvagent{} rather than from parallel computation across LocalAgents. As shown in Figure~\ref{fig:generation-cost}, our cumulative output is 55.7--73.1\% lower than that of the same-model agent baselines. We trace intermediate generation beyond the call-level averages in Table~\ref{tab:execution-stats} to understand this speed difference. Dependent decisions and observations execute sequentially, and each response requires token-by-token autoregressive decoding. Longer intermediate responses therefore accumulate latency throughout the analysis. LensWalk~\cite{li2026lenswalk} scans before inspecting, and each scan expands into visual calls over many segments that produce 66.3\% of its output. Video-RSI~\cite{luo2026videorsi} often answers after an overview with few actions, yet a minority of calls with more than 512 output tokens take over 80\% of call time. AgentCVR~\cite{qiu2026agentcvr} reasons in a central controller that also receives detailed observations, and its generation accumulates in both. Our observation instructions, observations, and LocalAgent reports focus on specific checks, with generally short responses at each stage. We achieve higher aggregate benchmark scores with less cumulative output than the same-model agent baselines, indicating higher output-token efficiency. Table~\ref{tab:execution-stats} shows that it views 109.9 frames per question on average with Qwen3.5, about 1/5 of the 512-frame budget of Single-pass inference, while scoring higher. This indicates that \mvagent{} locates question-relevant content more precisely.

\subsection{Further Analysis}
\label{sec:action-analysis}

\begin{figure}[!t]
\centering
\includegraphics[width=\columnwidth]{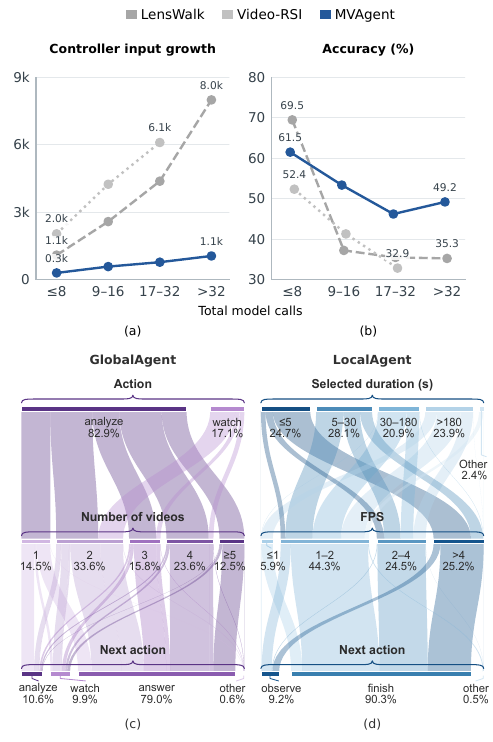}
\caption{\textbf{Execution patterns.} Three Qwen3.5 models. (a,b) Matched binary-scored questions grouped by total model calls; growth is the median first-to-peak input increase. (c,d) GlobalAgent acquisition and LocalAgent observation calls from all runs.}
\label{fig:trajectory-cost}
\label{fig:action-flows}
\end{figure}

\paragraph{Observation planning and context growth.}
\looseness=-1 Figure~\ref{fig:trajectory-cost}a shows that controller input grows faster with model calls for LensWalk~\cite{li2026lenswalk} and Video-RSI~\cite{luo2026videorsi} than for \mvagent{}. Both baselines use one controller to schedule observations across videos and make cross-video judgments, accumulating observations and intermediate decisions in a shared context. AgentCVR~\cite{qiu2026agentcvr} trajectories likewise show that its central controller retains previous tool responses from multiple videos when arranging subsequent observations and answering. In \mvagent{}, LocalAgents handle observations of their assigned videos, while GlobalAgent receives their reports. Continued per-video observation need not add all observation records to the global context. As shown in Figure~\ref{fig:trajectory-cost}b, slower input growth accompanies a gentler accuracy decline.

\paragraph{Global coordination and local autonomy.}
GlobalAgent can ask a LocalAgent for further evidence or use joint observation of selected clips, as shown in Figure~\ref{fig:action-flows}c,d. Among joint observations made after LocalAgent reports are available, 96.7\% involve only videos that have already returned reports. Follow-up joint observation usually checks relations between previously analyzed videos. We also measure video coverage, defined per question as the fraction of video length covered by the observed segments. Mean coverage is 97.5\% for \mvagent{}-9B and 66.0\% for AgentCVR-8B~\cite{qiu2026agentcvr}. Each LocalAgent selects the visual query, time intervals, and sampling rate for its video. Among LocalAgent requests containing at least two successful text-bearing observations, 67.0\% change the visual query, 33.2\% change the intervals, and 23.0\% change the sampling rate. These changes can co-occur. In AgentCVR, the central controller chooses each observation. In \mvagent{}, each LocalAgent adjusts its plan using its observations, decides whether to continue, and reports when it judges the request sufficiently addressed.

\paragraph{Temporal and cross-view reasoning.}
On the temporal-reasoning and procedural-ordering question group, mean accuracy across the three Qwen3.5 models exceeds the same-model Single-pass baseline by 33.6 points and same-model LensWalk~\cite{li2026lenswalk} and Video-RSI~\cite{luo2026videorsi} by 22.9--23.8 points. \mvagent{}-9B also exceeds AgentCVR-8B~\cite{qiu2026agentcvr} by 12.2 points on CVBench~\cite{zhu2026cvbench} temporal reasoning. On cross-view spatial reasoning, our three-model mean accuracy exceeds the stronger same-model baseline, Video-RSI, by 7.4 points. These tasks require different uses of evidence. Temporal reasoning requires establishing the events and states within each video before relating their order. GlobalAgent mainly reasons from per-video reports in these trajectories. Cross-view reasoning requires matching objects and positions across views, and its trajectories more often use joint observation. Per-video analysis provides observations of events, states, and objects. GlobalAgent combines the reports to establish relations between videos, while joint observation checks visual details across videos.

\paragraph{Limitations of evidence transfer and judgment.}
Errors can begin with missing or inaccurate observations of decisive details. Across the three Qwen3.5 configurations, 61.7\% of incorrect fine-grained discrimination trajectories do not accurately describe a key detail. Evidence can also be lost during synthesis. On open-ended questions, our rubric-item coverage is 10.3 points below the same-model Single-pass baseline. Trajectories show content needed for the answer being omitted from final answers even though it is present in LocalAgent reports, while some early errors persist through reports. Errors also arise when the needed evidence is present. Another 18.5\% of incorrect fine-grained discrimination trajectories contain evidence supporting the correct answer but reach an inconsistent judgment. On open-ended questions, correctness on the rubric items each method covers is 16.0 points below the same-model Single-pass baseline. AgentCVR~\cite{qiu2026agentcvr} remains stronger in some comparison and open-ended settings, and open-ended regression persists when \mvagent{} uses GPT-6-sol. In 87.7\% of incorrect fine-grained discrimination trajectories, the system uses only joint observation. Observation and judgment errors therefore also occur after joint observation, and compression in LocalAgent reports does not explain every error.

\subsection{Ablation Studies}
\looseness=-1
We compare model allocations and ablate components to examine the roles of perception, planning, and collaborative design, using trajectories to interpret the differences.
\begin{table}[t]
\centering
\caption{\textbf{Model allocation in \mvagent{}.} Scores: \%; $\Delta$ in pp vs. 9B planning model + 9B visual model.}
\label{tab:model-allocation}
\small
\setlength{\tabcolsep}{3pt}
\begin{tabular*}{0.9\columnwidth}{@{}c@{\hspace{14pt}}c@{\hspace{4pt}\vrule\hspace{4pt}\extracolsep{\fill}}ccc@{\hspace{5pt}}}
\toprule
Planning & Visual & CVBench & MVU-Eval & CrossVid\\
\midrule
9B & 9B & \ablbase{64.5} & \ablbase{50.5} & \ablbase{38.5}\\
9B & 27B & \abldelta{64.9}{featuregreen}{+0.4} & \abldelta{51.5}{featuregreen}{+0.9} & \abldelta{40.2}{featuregreen}{+1.7}\\
27B & 9B & \abldelta{68.8}{featuregreen}{+4.3} & \abldelta{55.3}{featuregreen}{+4.7} & \abldelta{40.4}{featuregreen}{+1.9}\\
27B & 27B & \abldelta{70.5}{featuregreen}{+6.0} & \abldelta{58.2}{featuregreen}{+7.6} & \abldelta{49.3}{featuregreen}{+10.7}\\
\bottomrule
\end{tabular*}
\end{table}

\Needspace{2\baselineskip}
\paragraph{Perception and planning.}
The preceding trajectories show errors in both perception and decisions about observation and evidence use, motivating a comparison of stronger perception and planning. Planning handles global decisions, per-video observation scheduling, and report generation. Perception performs visual observation through LocalAgent's \observe{} and GlobalAgent's joint observation. Table~\ref{tab:model-allocation} compares four Qwen3.5 model allocations. Strengthening planning yields larger overall gains than strengthening perception alone. From the all-27B configuration, reducing the planning model also causes a larger loss. The four configurations further show that the same visual-model upgrade yields greater gains with stronger planning, a trend shared by all three benchmarks. Thus, stronger planning yields larger system-level gains and better use of stronger perception.

\paragraph{Multi-agent organization.}
We remove LocalAgents and let GlobalAgent select a video and directly invoke the \observe{} action, appending the resulting observations to its global context. Accuracy falls below same-model Single-pass inference on all three benchmarks. We collect statistics on the reasoning trajectories, as shown in Table~\ref{tab:investigation-stats}. With a single controller, the system tends to stop exploring early, making fewer calls and observing a smaller share of each video. At the same time, raw observations enter the controller's context directly, so the global context grows faster and interferes with cross-video reasoning. Insufficient exploration and increased interference together cause a noticeable performance drop. The largest losses occur on tasks that need detailed observations of each video before cross-video comparison or ordering, because thinning per-video observation removes exactly the evidence those tasks require, and MVU-Eval~\cite{peng2025mvueval} counting is affected the most. Per-video organization therefore matters because it keeps observation depth per video while leaving the global context for relations between videos.

\paragraph{Joint observation.}
Per-video reports may omit visual details that determine cross-video relations. We keep LocalAgents but disable joint observation, leaving GlobalAgent to answer from per-video reports, to test whether joint observation recovers information that text reports cannot represent. Without it, performance drops on questions that require matching positions and objects across views, most on CrossVid~\cite{li2026crossvid}, which uses joint observation most often. This indicates that such information is hard to carry in reports. Unexpectedly, MVU-Eval~\cite{peng2025mvueval} improves. The trajectories show that, with \watch{} removed, some inappropriate joint observations that the system used to choose can no longer occur; these observations helped little on MVU-Eval questions but consumed decision rounds and brought irrelevant visual detail into the context. Answering from reports instead serves these questions better. Joint observation therefore matters when a question needs cross-video visual evidence, and deciding when to use it is itself a decision; the action-selection policy matters as much as the action itself.

\begin{table}[t]
\centering
\caption{\textbf{Component ablations.} Full 9B: complete \mvagent{}-9B. Scores: \%, $\Delta$ vs. Full 9B (pp).}
\label{tab:ablations}
\small
\setlength{\tabcolsep}{4pt}
\begin{tabularx}{\columnwidth}{@{}Xccc@{\hspace{5pt}}}
\toprule
Configuration & CVBench & MVU-Eval & CrossVid\\
\midrule
Full 9B & \ablbase{64.5} & \ablbase{50.5} & \ablbase{38.5}\\
\midrule
\hspace{1em}w/o LocalAgents & \abldelta{58.7}{featurered}{-5.8} & \abldelta{41.7}{featurered}{-8.9} & \abldelta{27.6}{featurered}{-10.9}\\
\hspace{1em}w/o joint observation & \abldelta{63.7}{featurered}{-0.8} & \abldelta{52.2}{featuregreen}{+1.6} & \abldelta{32.5}{featurered}{-6.0}\\
\hspace{1em}w/o cross-request history & \abldelta{64.4}{featurered}{-0.1} & \abldelta{50.5}{black!55}{0.0} & \abldelta{34.8}{featurered}{-3.7}\\
\bottomrule
\end{tabularx}
\end{table}

\begin{table}[t]
\centering
\caption{\textbf{Ablation statistics.} Per-question calls; coverage of video length observed; controller-input change (tokens).}
\label{tab:investigation-stats}
\small
\setlength{\tabcolsep}{3pt}
\renewcommand{\arraystretch}{1}
\begin{tabularx}{\columnwidth}{@{}l@{\hspace{6pt}}>{\hsize=0.85\hsize\centering\arraybackslash}X>{\hsize=0.85\hsize\centering\arraybackslash}X>{\hsize=1.15\hsize\centering\arraybackslash}X>{\hsize=1.15\hsize\centering\arraybackslash}X@{}}
\toprule
\multirow[c]{2}{*}[-2.5pt]{Config.} & \multicolumn{2}{c}{Calls} & \multirow[c]{2}{*}[-2.5pt]{Coverage} & \multirow[c]{2}{*}[-2.5pt]{$\Delta$ Ctx.}\\
\cmidrule(lr){2-3}
 & Plan & Visual & & \\
\midrule
Full 9B & 9.7 & 4.1 & 97.5\% & +473\\
\midrule
\hspace{1em}w/o LocalAgents & 3.0 & 3.2 & 86.7\% & +862\\
\bottomrule
\end{tabularx}
\end{table}

\paragraph{Cross-request history.}
To test whether later requests can continue from earlier observations, we clear LocalAgent history between requests while retaining within-request observations and reports already received by GlobalAgent. Overall, the change has little effect on performance. Statistics on Full 9B trajectories show that few questions need a second call; only 12.1\% of CrossVid~\cite{li2026crossvid} questions call the same LocalAgent again, versus 0.6\% on CVBench and 4.1\% on MVU-Eval~\cite{peng2025mvueval}, so CrossVid is where the change matters most. Retaining history nevertheless helps action selection. Removing it increases observations with identical video, interval, and sampling rate by 47.2\% and cross-request repeats by 120.7\%, showing that history lets later requests reuse earlier observations and avoid redundant viewing.


\FloatBarrier
\section{Conclusion and Limitations}
\label{sec:conclusion}
We presented \mvagent{}, a training-free framework that coordinates per-video analysis for cross-video reasoning. Across three benchmarks, it outperforms Single-pass inference in all settings and prior state-of-the-art agents in nearly all, while using fewer output tokens and running faster than these agents. Memory management, the way agents exchange information, and more deliberate exploration strategies all remain open. Reports compress local observations, so details that open-ended answers need can be lost before synthesis, and a better exchange format could preserve these details.

\clearpage
\let\arxivparagraph\paragraph\mvagentrestoreparagraphspacing
\clearpage
\label{paper:references-start}
{\small
\bibliographystyle{ieeenat_fullname}

}
\clearpage
\let\paragraph\arxivparagraph
\maketitlesupplementary
\raggedbottom
\renewcommand{\floatpagefraction}{.7}
\renewcommand{\dblfloatpagefraction}{.7}
\appendix
\setcounter{table}{0}\setcounter{figure}{0}
\renewcommand{\theHtable}{supp.\arabic{table}}
\renewcommand{\theHfigure}{supp.\arabic{figure}}
\renewcommand{\thetable}{S\arabic{table}}
\renewcommand{\thefigure}{S\arabic{figure}}

\section{Additional Related Work}
\label{app:related-work}
\subsection{Context organization in agent systems}
Agents following ReAct~\cite{yao2023react} interleave reasoning and actions, using prior observations to guide subsequent decisions. As observations accumulate, agents need to manage context and memory for later reasoning. MemGPT~\cite{packer2023memgpt} moves information between memory tiers, while ReadAgent~\cite{lee2024readagent} compresses document episodes into gist memories and revisits original passages when more detail is needed. Mem0~\cite{chhikara2025mem0} uses vector embeddings to retrieve semantically related records.

Other methods distribute inputs and memories across local contexts. LONGAGENT~\cite{zhao2024longagent} and Chain of Agents~\cite{zhang2024chainofagents} assign document chunks to workers. LONGAGENT coordinates members through a leader over multiple rounds, while Chain of Agents passes information between workers before a manager integrates their outputs. HMARS~\cite{li2026hmars} assigns bounded memory regions to sub-agents and aggregates local memory cards hierarchically. Recursive Language Models~\cite{zhang2025rlm} keep inputs in an external environment and recursively call a model on selected snippets.

\subsection{Single-video understanding}
Single-video benchmarks assess perception and reasoning across temporal scales. Video-MME~\cite{fu2025videomme} evaluates understanding across domains and video durations, while EgoSchema~\cite{mangalam2023egoschema} emphasizes egocentric activities. Long-video benchmarks cover complementary demands, including holistic and detailed understanding in MLVU~\cite{zhou2025mlvu}, referring reasoning over video and subtitles in LongVideoBench~\cite{wu2024longvideobench}, and hour-long video comprehension in LVBench~\cite{wang2025lvbench}. Charades-STA~\cite{gao2017tall} evaluates temporal grounding of natural-language queries.

Video context construction uses memory, compression, and question-dependent selection. MovieChat~\cite{song2024moviechat} and MA-LMM~\cite{he2024mallm} retain earlier visual information in memory, while FlexMem~\cite{chen2026flexmem} stores and selectively reads compressed visual key--value caches. LongVU~\cite{shen2024longvu} and VideoChat-Flash~\cite{li2026videochatflash} reduce spatial and temporal redundancy, while Tempo~\cite{fei2026tempo} uses a small vision-language model for question-conditioned compression and adaptive token allocation. LongVILA~\cite{chen2025longvila} extends the supported input length through long-context training. Flexible Frame Selection~\cite{buch2025ffs} learns which and how many frames to select, while VideoTree~\cite{wang2025videotree} builds a query-adaptive evidence hierarchy.

Video agents use memory retrieval and further observation to gather evidence iteratively. Fan et al.'s VideoAgent~\cite{fan2024memoryvideoagent} combines temporal event memory with object records stored in an SQL database. DrVideo~\cite{ma2025drvideo} and Deep Video Discovery~\cite{zhang2025dvd} organize retrieval around video documents and databases, while VideoARM~\cite{yin2026videoarm} and Visual Agentic Memory~\cite{li2026vam} use hierarchical memories. Wang et al.'s VideoAgent~\cite{wang2024videoagent} and VCA~\cite{yang2025vca} adapt retrieval or segment exploration to current information needs. AVP~\cite{wang2026avp} iterates planning, targeted observation, and reflection; LensWalk~\cite{li2026lenswalk} gives the reasoner control over temporal scope and sampling density. Observation procedures can also be improved through policy learning or offline program revision. EVA~\cite{zhang2026eva} learns observation policies, while Video-o3~\cite{zeng2026videoo3} and Weaver~\cite{shi2026weaver} learn to interleave visual tool calls with reasoning. Video-RSI~\cite{luo2026videorsi} revises its reasoning harness offline using execution feedback with frozen weights.

Functional specialization and collaboration organize video reasoning across models or roles. LVAgent~\cite{chen2025lvagent} and A4VL~\cite{xu2026a4vl} refine answers through multi-agent interaction, while Symphony~\cite{yan2026symphony} and VideoMultiAgents~\cite{kugo2025videomultiagents} coordinate specialized perception and analysis roles. VideoMind~\cite{liu2026videomind} implements planning, grounding, verification, and answering through LoRA adapters on a shared backbone.

\subsection{Multi-video understanding}
Multi-video benchmarks assess relationships and evidence across sources. CVBench~\cite{zhu2026cvbench} evaluates entity and event association and knowledge integration across videos. MVU-Eval~\cite{peng2025mvueval} covers abilities such as counting, comparison, and temporal reasoning, while CrossVid~\cite{li2026crossvid} includes comparative, temporal, and multi-view reasoning. MVX-Bench~\cite{zhang2026sama} tests cross-video identity matching and structured reasoning.

Complementary benchmarks examine other aspects of multi-video understanding. MVPBench~\cite{bai2026mvpbench} examines multi-video perception, VideoSetDiff~\cite{qiu2025videosetdiff} focuses on fine-grained similarities and differences among videos, and $M^3$-Verse~\cite{wei2026m3verse} evaluates scene changes across paired videos. Related multi-view image benchmarks such as All-Angles Bench~\cite{yeh2026allangles} assess spatial relationships and camera poses across views of the same scene.

Models for multiple visual inputs provide a foundation for cross-video reasoning. Mantis~\cite{jiang2024mantis} and LLaVA-Interleave~\cite{li2025llavainterleave} support reasoning over multiple images or interleaved visual inputs. Dedicated comparison methods, including StepDiff~\cite{nagarajan2024stepdiff} and VidDiff~\cite{burgess2025viddiff}, identify procedural or action differences between videos.

Retrieval-augmented generation selects evidence from larger video collections. Jeong et al.'s VideoRAG~\cite{jeong2025videorag} combines retrieved visual and textual content, while Ren et al.'s VideoRAG~\cite{ren2025videorag} uses graph-based textual knowledge and multimodal retrieval.

Multi-video systems organize evidence through role decomposition and coordinated acquisition. MAGNET~\cite{chowdhury2025magnet} extracts evidence through video-specific roles before centralized aggregation, while VideoForest~\cite{meng2025videoforest} uses hierarchical reasoning over person-centered evidence. Other systems iteratively acquire evidence under a central planner. SAMA~\cite{zhang2026sama} combines reusable task-specific Skills, multimodal tools, external verification, and video-indexed evidence. AgentCVR~\cite{qiu2026agentcvr} coordinates modality-specific agents through a trained Master Agent. AdSeek~\cite{zhao2026adseek} learns multi-round acquisition across videos, intervals, modalities, and visual regions for e-commerce comparisons.

\section{Implementation Details}
\label{app:implementation}

\subsection{Execution and inputs}
\label{app:execution-settings}
\label{app:memory-flow}

Table~\ref{app:settings} summarizes the base configuration.

\begin{table}[tbp]
\centering
\caption{\textbf{Base local configuration.} Global rounds are capped per question, local steps per request, and output tokens per call.}
\label{app:settings}
\small
\setlength{\tabcolsep}{3pt}
\renewcommand{\arraystretch}{1.18}
\setlength{\aboverulesep}{1pt}
\setlength{\belowrulesep}{1pt}
\begin{tabularx}{\columnwidth}{>{\raggedright\arraybackslash}Xc|c}
\toprule
\rowcolor{agentblue!12}
\textbf{Setting} & \multicolumn{2}{c}{\textbf{Value}}\\
\midrule
\rowcolor{black!3}
Global rounds & \multicolumn{2}{c}{5}\\
Local steps & \multicolumn{2}{c}{5}\\
\rowcolor{black!3}
Output tokens / call & \multicolumn{2}{c}{2,048}\\
Max. frames / visual call & \multicolumn{2}{c}{512}\\
\rowcolor{black!3}
Thinking & Qwen: \texttt{off} & GPT: \texttt{low}\\
\bottomrule
\end{tabularx}
\end{table}

For locally deployed Qwen models, video clips are sent to vLLM through the \texttt{video\_url} field as Base64-encoded video data URLs, with decoding and frame sampling handled on the server. For GPT-6-sol, frames are sampled before each visual request and sent as Base64-encoded image data URLs.

\subsection{Validation and fallback}
\label{app:validation-fallback}
Action schemas are generated from the available actions, source IDs, durations, and field limits. The runtime validates JSON structure and action parameters before execution. If a planning or reporting response fails parsing or schema validation, the same model receives at most one repair request with the original request, required schema, validation error, and prior response. Invalid decisions and observation failures are recorded as feedback and consume the current step; subsequent decisions use the remaining budget. Invalid visual responses are returned as errors without a separate repair call.

Table~\ref{app:action-limits} summarizes the observation limits.

\begin{table}[tbp]
\centering
\caption{\textbf{Limits per observation action.}}
\label{app:action-limits}
\small
\setlength{\tabcolsep}{3pt}
\renewcommand{\arraystretch}{1.18}
\setlength{\aboverulesep}{1pt}
\setlength{\belowrulesep}{1pt}
\begin{tabularx}{\columnwidth}{l>{\raggedright\arraybackslash}Xc}
\toprule
\rowcolor{agentblue!12}
\textbf{Action} & \textbf{Parameter} & \textbf{Value}\\
\midrule
\rowcolor{black!3}
 & Number of temporal ranges & 8\\
\rowcolor{black!3}
\multirow{-2}{*}{\observe{}} & Planned sampling rate & 0.1--25.0 fps\\
\midrule
\watch{} & Number of videos & 2--4\\
\bottomrule
\end{tabularx}
\end{table}

The planning model generates ordinary and terminal \finish{} reports. At the local step limit, further \observe{} actions stop. If retained observations exist, an additional text-only request produces the report, using no system message and a schema containing only \texttt{summary}; ordinary \finish{} uses the action-and-parameters schema. If no observations exist or this summary fails, the LocalAgent returns an empty report with error feedback.

At the global round limit, further evidence collection stops and GlobalAgent must answer from its accumulated global history. It receives an additional decision request using the same decision template and system message, with available actions restricted to \answer{}. If this terminal decision fails, the system returns an empty answer and records the failure. Repair and terminal requests are additional model calls beyond the ordinary decision steps.

\subsection{Evaluation settings}
\label{app:evaluation-settings}

\paragraph{Benchmark sampling.}
\begin{figure}[!tbp]
\centering
\includegraphics[width=\columnwidth,trim=0 7bp 0 0,clip]{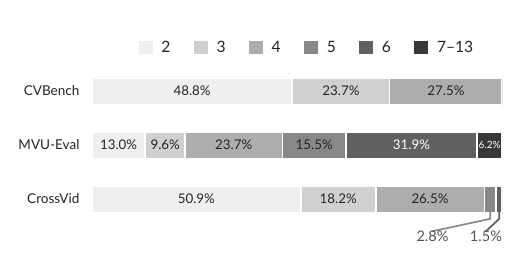}
\par\vspace{4pt}
\includegraphics[width=\columnwidth]{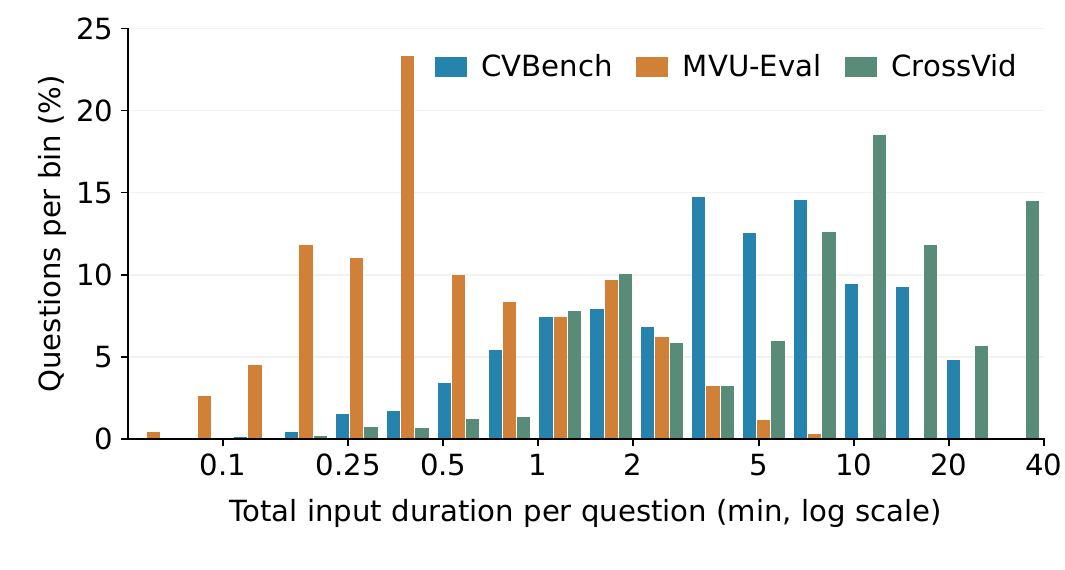}
\caption{\textbf{Input-video counts (top) and total durations (bottom) per question.} Duration bins are uniform on a log scale; CrossVid uses the 2,500-question subset.}
\label{app:input-distributions}
\end{figure}
To improve evaluation efficiency, we use a stratified random subset of 2,500 CrossVid~\cite{li2026crossvid} questions. Even after sampling, this subset still contains substantially more questions than CVBench~\cite{zhu2026cvbench} (1,000 questions) and MVU-Eval~\cite{peng2025mvueval} (1,824 questions), providing a useful basis for comparing multi-video understanding methods. Figure~\ref{app:input-distributions} shows the input-video count and total duration per question. Total duration sums all video inputs, with question percentages computed within each benchmark. Table~\ref{app:crossvid-coverage} gives the CrossVid sampling breakdown. Cmp, Tmp, View, and Free denote Comparative Analysis, Temporal Understanding, Multi-view Reasoning, and Free-form QA, respectively. We construct the subset as follows.
\begin{enumerate}\tightitems
\item \textbf{Set task quotas.} Require at least 100 questions per task and approximately match sampling fractions across tasks.
\item \textbf{Stratify each task.} Form strata from input-video count, total presented-video duration, question-text length, and whether annotated temporal segments outnumber input videos. Set duration boundaries at the task-specific 25th, 50th, 75th, and 90th percentiles, merging duplicate boundaries, and split question-text length at its median.
\item \textbf{Allocate within each task.} Give each nonempty stratum at least one question and allocate the remainder approximately in proportion to stratum size.
\item \textbf{Sample and freeze.} Sample without replacement, without using answer labels or model predictions, and freeze the selected IDs across model configurations.
\end{enumerate}

\begin{table}[tbp]
\centering
\caption{\textbf{CrossVid sampling rates.} Rate denotes the sampling rate (Sampled/Original).}
\label{app:crossvid-coverage}
\small
\setlength{\tabcolsep}{3pt}
\renewcommand{\arraystretch}{1.18}
\setlength{\aboverulesep}{1pt}
\setlength{\belowrulesep}{1pt}
\begin{tabularx}{\columnwidth}{@{}>{\centering\arraybackslash}p{1.6cm}>{\centering\arraybackslash}p{1cm}*{3}{>{\centering\arraybackslash}X}@{}}
\toprule
\rowcolor{agentblue!12}
\textbf{Dimension} & \textbf{Task} & \textbf{Original} & \textbf{Sampled} & \textbf{Rate}\\
\midrule
\rowcolor{black!3}
 & BU & 848 & 232 & 27.36\%\\
\rowcolor{black!3}
 & NC & 1,221 & 334 & 27.35\%\\
\rowcolor{black!3}
 & CC & 798 & 218 & 27.32\%\\
\rowcolor{black!3}
\multirow{-4}{=}{\centering Cmp} & PEA & 953 & 261 & 27.39\%\\
\midrule
 & PI & 251 & 100 & 39.84\%\\
 & FSA & 2,248 & 616 & 27.40\%\\
\multirow{-3}{=}{\centering Tmp} & PSS & 664 & 182 & 27.41\%\\
\midrule
\rowcolor{black!3}
 & MSR & 594 & 163 & 27.44\%\\
\rowcolor{black!3}
\multirow{-2}{=}{\centering View} & MOC & 566 & 155 & 27.39\%\\
\midrule
Free & CCQA & 872 & 239 & 27.41\%\\
\midrule
\rowcolor{agentblue!8}
\multicolumn{2}{@{}c}{\textbf{Total}} & \textbf{9,015} & \textbf{2,500} & \textbf{27.73\%}\\
\bottomrule
\end{tabularx}
\end{table}

PI's higher sampling rate follows from the minimum question quota; the other tasks have closely matched rates. Scores describe this fixed research subset rather than the full CrossVid benchmark.

\paragraph{Metrics and judging.}
We follow the scoring scripts and metric definitions in the official benchmark repositories. For CVBench and MVU-Eval, overall accuracy is
\begin{equation}
\mathrm{Acc}_{D}=\frac{100}{n_D}\sum_{i=1}^{n_D}\mathbf{1}[\hat y_i=y_i],
\end{equation}
where $D$ is either benchmark, $n_D$ is its number of evaluated questions, and $\hat y_i$ and $y_i$ are the parsed prediction and reference answer. Category scores use the same formula over the corresponding questions.

CrossVid reports an unweighted mean over its ten task scores,
\begin{equation}
\begin{aligned}
S_t&=\frac{100}{n_t}\sum_{i=1}^{n_t}s_{t,i}, \quad t\ne\mathrm{CCQA},\\
S_{\mathrm{CrossVid}}&=\frac{1}{10}\sum_{t=1}^{10}S_t,
\end{aligned}
\end{equation}
where $n_t$ is the number of evaluated questions in task $t$. For discrete-answer tasks, $s_{t,i}$ is 1 only when the complete parsed answer matches the reference, and 0 otherwise. For FSA, the per-question score is temporal interval IoU,
\begin{equation}
s_{\mathrm{FSA},i}=\frac{|\hat I_i\cap I_i|}{|\hat I_i\cup I_i|},
\end{equation}
where $\hat I_i$ and $I_i$ are the predicted and reference intervals, and $|\cdot|$ denotes duration.

For CCQA, the official coverage/correctness rubric aggregates over scoring points,
\begin{equation}
\begin{aligned}
S_{\mathrm{CCQA}}&=100\frac{\sum_i(C_i-E_i/2)}{\sum_i K_i}\\
&=100\frac{\sum_i(C_i+R_i)}{2\sum_i K_i},
\end{aligned}
\end{equation}
where $K_i$ is the number of rubric points for question $i$, $C_i$ counts covered points, $E_i$ counts covered-but-incorrect points, and $R_i=C_i-E_i$ counts correctly covered points. Thus, covered but incorrect points receive half credit, and uncovered points receive zero. CCQA is weighted by rubric-point count rather than averaged equally across questions. Saved CCQA answers are scored offline with DeepSeek-V4.1-Flash~\cite{deepseek2026flashdocs} using the official judging prompt.

The fixed judging prompt is reproduced below.
\begin{promptbox}{CCQA open-ended scoring}{prompts/ccqa_judge_prompt.tex}
\begin{Verbatim}
You are asked to score the output of a model, given the following information:
- Question: {QUESTION}
- Standard Answer: {ANSWER}
- Scoring Points: {POINTS}
- Model's Output: {OUTPUT}

Please perform the following two-part scoring:
Part 1: Coverage of Scoring Points
- For each scoring point, determine whether it is covered by the Model's Output.
- Mark as covered (true) **only if** the scoring point is addressed **explicitly, independently, and clearly**.
- If the mention is vague, partial, or ambiguous, consider it **not covered**.

Part 2: Accuracy of Details
- For each covered scoring point, compare the details in the Model's Output to the Standard Answer.
- Mark as correct (true) **only if** the details are **fully accurate and consistent** with the Standard Answer, without any error, omission, or ambiguity.
- If the answer is partially correct, too broad/narrow, or not strictly consistent, mark it as **not correct** (false).
- For scoring points not covered, mark as incorrect.

Format your answer in a json format as follows:
{
    "coverage": [true, false, true, ...],
    "correctness": [true, false, false, ...]
}
The length of 'coverage' and 'correctness' lists should match the number of scoring points.
Wrap the json output within <score></score> tags.

Your answer:
\end{Verbatim}

\end{promptbox}

\paragraph{Baseline budgets and timing.}
We follow the inference procedures in the official baseline repositories, with the input and answer-format adaptations described in the main paper. LensWalk~\cite{li2026lenswalk} allows at most five tool calls per question. Video-RSI~\cite{luo2026videorsi} uses budgets of 11 steps, 64 service calls, and 256 cumulative frame exposures per question.

Local runtime is measured on NVIDIA H100 GPUs. Time denotes mean saved execution seconds per question, including failed executions and excluding separately recorded preparation, archival, and offline judging. GPT-6-sol runtime is unmeasured because remote API latency includes network and serving delays.

\subsection{Ablation configurations}
\label{app:ablation-details}
The component ablations use Qwen3.5-9B for the remaining planning and visual roles on the evaluation panels in Sec.~\ref{app:evaluation-settings}.

\paragraph{Without LocalAgents.}
We remove the per-video planners and their private histories. GlobalAgent's role and action prompts are updated to directly invoke \observe{} with a selected video, query, interval, and FPS, and the returned observations are appended to its global history. The model-call limit is set to 16 per question.

\paragraph{Without joint observation.}
We disable \watch{} and remove its action description and joint-inspection guidance from the global prompt.

\paragraph{Without cross-request history.}
We reset a LocalAgent's private history when it receives a new request.

\section{Detailed Experimental Results}
\label{app:detailed-results}
\setlength{\dblfloatsep}{6pt}

\subsection{Task-level Results}
\label{app:task-level-results}

\begin{table*}[p]
\centering
\caption{\textbf{Detailed CVBench results~\cite{zhu2026cvbench}.} Average is question-weighted. Prefixes M., J., and C. denote multi-view/video, joint-video, and cross-video. SU, scene understanding; TR, temporal reasoning; SN, spatial navigation; VDC, video difference captioning; CR, counterfactual reasoning; S, summarization; PT, procedural transfer; OR, object recognition; AR, attribute recognition; C, counting; EM, entity matching; AD, anomaly detection; SR, scene recognition; KAR, key-action recognition; ER, event retrieval.}
\label{app:cvbench-detailed}
\footnotesize
\renewcommand{\arraystretch}{1.24}
\setlength{\aboverulesep}{1pt}
\setlength{\belowrulesep}{1pt}
\setlength{\cmidrulesep}{1pt}
\centering
\footnotesize
\setlength{\tabcolsep}{1.1pt}
\begin{tabularx}{\textwidth}{>{\raggedright\arraybackslash}p{68pt}!{\tikz[remember picture,overlay]{\coordinate (detail-reasoning-v1);}}>{\centering\arraybackslash}X!{\tikz[remember picture,overlay]{\coordinate (detail-reasoning-v2);}}>{\centering\arraybackslash}X>{\centering\arraybackslash}X>{\centering\arraybackslash}X>{\centering\arraybackslash}X>{\centering\arraybackslash}X>{\centering\arraybackslash}X>{\centering\arraybackslash}X}
\toprule
\noalign{\hbox{\tikz[remember picture,overlay]{\coordinate (detail-reasoning-top);}}}
\multirow{2}{=}[-0.7ex]{\centering\textbf{Method}} & \multicolumn{8}{c}{\textbf{Cross-video complex reasoning}}\\
\cmidrule{2-9}
\noalign{\hbox{\tikz[remember picture,overlay]{\coordinate (detail-reasoning-subtop);}}}
 & \textbf{Average} & \textbf{M. SU} & \textbf{M. TR} & \textbf{J. SN} & \textbf{VDC} & \textbf{C. CR} & \textbf{J. S} & \textbf{C. PT}\\
\noalign{\hbox{\tikz[remember picture,overlay]{\coordinate (detail-reasoning-header-bottom);}}}
\midrule
\rowcolor{black!8}
\multicolumn{9}{l}{\textit{AgentCVR-8B}~\cite{qiu2026agentcvr}}\\
\noalign{\hbox{\tikz[remember picture,overlay]{\coordinate (detail-reasoning-g0-top);}}}
\hspace{4pt}AgentCVR & 55.9 & 78.2 & 50.0 & 38.1 & 32.7 & 58.8 & 71.2 & 62.0\\
\noalign{\hbox{\tikz[remember picture,overlay]{\coordinate (detail-reasoning-g0-bottom);}}}
\midrule
\rowcolor{black!8}
\multicolumn{9}{l}{\textit{Qwen3.5-9B}}\\
\noalign{\hbox{\tikz[remember picture,overlay]{\coordinate (detail-reasoning-g1-top);}}}
\hspace{4pt}\textcolor{black!55}{\textit{Single-pass}} & \textcolor{black!55}{\textbf{62.3}} & \textcolor{black!55}{\textbf{83.6}} & \textcolor{black!55}{37.8} & \textcolor{black!55}{\textbf{45.2}} & \textcolor{black!55}{\textbf{69.1}} & \textcolor{black!55}{\underline{62.7}} & \textcolor{black!55}{\textbf{78.8}} & \textcolor{black!55}{\textbf{64.0}}\\
\cmidrule(lr){1-9}
\hspace{4pt}LensWalk~\cite{li2026lenswalk} & 55.1 & 78.2 & \underline{50.0} & 35.7 & 49.1 & 52.9 & 73.1 & 44.0\\
\hspace{4pt}Video-RSI~\cite{luo2026videorsi} & \underline{56.5} & 74.5 & 36.5 & \underline{42.9} & 52.7 & \textbf{66.7} & \underline{76.9} & 50.0\\
\specialrule{\lightrulewidth}{0pt}{0pt}
\rowcolor{mvagentrow}
\hspace{4pt}\textbf{MVAgent} & \textbf{62.3} & \underline{81.8} & \textbf{62.2} & \textbf{45.2} & \underline{54.5} & 58.8 & 75.0 & \underline{54.0}\\
\noalign{\hbox{\tikz[remember picture,overlay]{\coordinate (detail-reasoning-g1-bottom);}}}
\midrule
\rowcolor{black!8}
\multicolumn{9}{l}{\textit{Qwen3.5-35B-A3B}}\\
\noalign{\hbox{\tikz[remember picture,overlay]{\coordinate (detail-reasoning-g2-top);}}}
\hspace{4pt}\textcolor{black!55}{\textit{Single-pass}} & \textcolor{black!55}{\textbf{65.4}} & \textcolor{black!55}{\underline{83.6}} & \textcolor{black!55}{39.2} & \textcolor{black!55}{\underline{47.6}} & \textcolor{black!55}{\textbf{70.9}} & \textcolor{black!55}{\textbf{72.5}} & \textcolor{black!55}{\textbf{75.0}} & \textcolor{black!55}{\textbf{76.0}}\\
\cmidrule(lr){1-9}
\hspace{4pt}LensWalk & 53.8 & 63.6 & \underline{48.6} & 28.6 & 50.9 & \underline{70.6} & 53.8 & 58.0\\
\hspace{4pt}Video-RSI & 59.4 & 78.2 & 41.9 & \textbf{50.0} & \underline{56.4} & 62.7 & 71.2 & 60.0\\
\specialrule{\lightrulewidth}{0pt}{0pt}
\rowcolor{mvagentrow}
\hspace{4pt}\textbf{MVAgent} & \underline{63.9} & \textbf{87.3} & \textbf{64.9} & 40.5 & 47.3 & 58.8 & \underline{73.1} & \underline{70.0}\\
\noalign{\hbox{\tikz[remember picture,overlay]{\coordinate (detail-reasoning-g2-bottom);}}}
\midrule
\rowcolor{black!8}
\multicolumn{9}{l}{\textit{Qwen3.5-27B}}\\
\noalign{\hbox{\tikz[remember picture,overlay]{\coordinate (detail-reasoning-g3-top);}}}
\hspace{4pt}\textcolor{black!55}{\textit{Single-pass}} & \textcolor{black!55}{64.9} & \textcolor{black!55}{\underline{85.5}} & \textcolor{black!55}{36.5} & \textcolor{black!55}{\textbf{57.1}} & \textcolor{black!55}{\underline{65.5}} & \textcolor{black!55}{\underline{64.7}} & \textcolor{black!55}{\textbf{80.8}} & \textcolor{black!55}{\textbf{74.0}}\\
\cmidrule(lr){1-9}
\hspace{4pt}LensWalk & 63.1 & 76.4 & \underline{56.8} & 40.5 & 60.0 & 60.8 & 71.2 & \textbf{74.0}\\
\hspace{4pt}Video-RSI & \underline{65.4} & 83.6 & 55.4 & 42.9 & \textbf{67.3} & \underline{64.7} & \underline{78.8} & 64.0\\
\specialrule{\lightrulewidth}{0pt}{0pt}
\rowcolor{mvagentrow}
\hspace{4pt}\textbf{MVAgent} & \textbf{70.2} & \textbf{90.9} & \textbf{71.6} & \underline{45.2} & 63.6 & \textbf{66.7} & \underline{78.8} & \underline{68.0}\\
\noalign{\hbox{\tikz[remember picture,overlay]{\coordinate (detail-reasoning-g3-bottom);}}}
\midrule
\rowcolor{black!8}
\multicolumn{9}{l}{\textit{GPT-6-sol}}\\
\noalign{\hbox{\tikz[remember picture,overlay]{\coordinate (detail-reasoning-g4-top);}}}
\hspace{4pt}\textcolor{black!55}{\textit{Single-pass}} & \textcolor{black!55}{\underline{74.4}} & \textcolor{black!55}{\textbf{85.5}} & \textcolor{black!55}{\underline{67.6}} & \textcolor{black!55}{\underline{59.5}} & \textcolor{black!55}{\underline{72.7}} & \textcolor{black!55}{\textbf{74.5}} & \textcolor{black!55}{\textbf{84.6}} & \textcolor{black!55}{\textbf{76.0}}\\
\specialrule{\lightrulewidth}{0pt}{0pt}
\rowcolor{mvagentrow}
\hspace{4pt}\textbf{MVAgent} & \textbf{75.7} & \underline{81.8} & \textbf{70.3} & \textbf{85.7} & \textbf{81.8} & \underline{64.7} & \underline{76.9} & \underline{72.0}\\
\noalign{\hbox{\tikz[remember picture,overlay]{\coordinate (detail-reasoning-g4-bottom);}}}
\bottomrule
\end{tabularx}
\tikz[remember picture,overlay]{
\draw[black,line width=1.2pt] (detail-reasoning-v1 |- detail-reasoning-top) -- (detail-reasoning-v1 |- detail-reasoning-header-bottom);
\draw[black,line width=1.2pt] (detail-reasoning-v1 |- detail-reasoning-g0-top) -- (detail-reasoning-v1 |- detail-reasoning-g0-bottom);
\draw[black,line width=1.2pt] (detail-reasoning-v1 |- detail-reasoning-g1-top) -- (detail-reasoning-v1 |- detail-reasoning-g1-bottom);
\draw[black,line width=1.2pt] (detail-reasoning-v1 |- detail-reasoning-g2-top) -- (detail-reasoning-v1 |- detail-reasoning-g2-bottom);
\draw[black,line width=1.2pt] (detail-reasoning-v1 |- detail-reasoning-g3-top) -- (detail-reasoning-v1 |- detail-reasoning-g3-bottom);
\draw[black,line width=1.2pt] (detail-reasoning-v1 |- detail-reasoning-g4-top) -- (detail-reasoning-v1 |- detail-reasoning-g4-bottom);
\draw[black,line width=1.2pt] (detail-reasoning-v2 |- detail-reasoning-subtop) -- (detail-reasoning-v2 |- detail-reasoning-header-bottom);
\draw[black,line width=1.2pt] (detail-reasoning-v2 |- detail-reasoning-g0-top) -- (detail-reasoning-v2 |- detail-reasoning-g0-bottom);
\draw[black,line width=1.2pt] (detail-reasoning-v2 |- detail-reasoning-g1-top) -- (detail-reasoning-v2 |- detail-reasoning-g1-bottom);
\draw[black,line width=1.2pt] (detail-reasoning-v2 |- detail-reasoning-g2-top) -- (detail-reasoning-v2 |- detail-reasoning-g2-bottom);
\draw[black,line width=1.2pt] (detail-reasoning-v2 |- detail-reasoning-g3-top) -- (detail-reasoning-v2 |- detail-reasoning-g3-bottom);
\draw[black,line width=1.2pt] (detail-reasoning-v2 |- detail-reasoning-g4-top) -- (detail-reasoning-v2 |- detail-reasoning-g4-bottom);
}

\par\vspace{6pt}
\centering
\footnotesize
\setlength{\tabcolsep}{1.1pt}
\begin{tabularx}{\textwidth}{>{\raggedright\arraybackslash}p{68pt}!{\tikz[remember picture,overlay]{\coordinate (detail-association-v1);}}>{\centering\arraybackslash}X!{\tikz[remember picture,overlay]{\coordinate (detail-association-v2);}}*{4}{>{\centering\arraybackslash}X}!{\tikz[remember picture,overlay]{\coordinate (detail-association-v6);}}>{\centering\arraybackslash}X!{\tikz[remember picture,overlay]{\coordinate (detail-association-v7);}}*{4}{>{\centering\arraybackslash}X}}
\toprule
\noalign{\hbox{\tikz[remember picture,overlay]{\coordinate (detail-association-top);}}}
\multirow{2}{=}[-0.7ex]{\centering\textbf{Method}} & \multicolumn{5}{c}{\textbf{Cross-video object association}} & \multicolumn{5}{c}{\textbf{Cross-video event association}}\\
\cmidrule{2-11}
\noalign{\hbox{\tikz[remember picture,overlay]{\coordinate (detail-association-subtop);}}}
 & \textbf{Average} & \textbf{C. OR} & \textbf{M. AR} & \textbf{J. C} & \textbf{C. EM} & \textbf{Average} & \textbf{C. AD} & \textbf{C. SR} & \textbf{M. KAR} & \textbf{C. ER}\\
\noalign{\hbox{\tikz[remember picture,overlay]{\coordinate (detail-association-header-bottom);}}}
\midrule
\rowcolor{black!8}
\multicolumn{11}{l}{\textit{AgentCVR-8B}~\cite{qiu2026agentcvr}}\\
\noalign{\hbox{\tikz[remember picture,overlay]{\coordinate (detail-association-g0-top);}}}
\hspace{4pt}AgentCVR & 56.5 & 55.9 & 67.4 & 35.0 & 67.6 & 58.1 & 61.9 & 57.7 & 56.8 & 55.1\\
\noalign{\hbox{\tikz[remember picture,overlay]{\coordinate (detail-association-g0-bottom);}}}
\midrule
\rowcolor{black!8}
\multicolumn{11}{l}{\textit{Qwen3.5-9B}}\\
\noalign{\hbox{\tikz[remember picture,overlay]{\coordinate (detail-association-g1-top);}}}
\hspace{4pt}\textcolor{black!55}{\textit{Single-pass}} & \textcolor{black!55}{\underline{60.9}} & \textcolor{black!55}{\underline{63.2}} & \textcolor{black!55}{\underline{71.7}} & \textcolor{black!55}{\textbf{45.0}} & \textcolor{black!55}{64.9} & \textcolor{black!55}{\underline{62.7}} & \textcolor{black!55}{\underline{53.6}} & \textcolor{black!55}{63.8} & \textcolor{black!55}{\underline{67.0}} & \textcolor{black!55}{\textbf{67.3}}\\
\cmidrule(lr){1-11}
\hspace{4pt}LensWalk~\cite{li2026lenswalk} & 55.2 & 61.8 & \underline{71.7} & 33.3 & 56.8 & 58.1 & 51.2 & 60.4 & 58.0 & 63.3\\
\hspace{4pt}Video-RSI~\cite{luo2026videorsi} & 57.7 & 55.9 & 69.6 & 38.3 & \underline{67.6} & 59.7 & 48.8 & \underline{67.8} & 61.4 & 51.0\\
\specialrule{\lightrulewidth}{0pt}{0pt}
\rowcolor{mvagentrow}
\hspace{4pt}\textbf{MVAgent} & \textbf{64.9} & \textbf{70.6} & \textbf{76.1} & \underline{43.3} & \textbf{70.3} & \textbf{66.5} & \textbf{57.1} & \textbf{69.1} & \textbf{71.6} & \underline{65.3}\\
\noalign{\hbox{\tikz[remember picture,overlay]{\coordinate (detail-association-g1-bottom);}}}
\midrule
\rowcolor{black!8}
\multicolumn{11}{l}{\textit{Qwen3.5-35B-A3B}}\\
\noalign{\hbox{\tikz[remember picture,overlay]{\coordinate (detail-association-g2-top);}}}
\hspace{4pt}\textcolor{black!55}{\textit{Single-pass}} & \textcolor{black!55}{62.9} & \textcolor{black!55}{66.2} & \textcolor{black!55}{\underline{76.1}} & \textcolor{black!55}{\underline{46.7}} & \textcolor{black!55}{64.9} & \textcolor{black!55}{63.8} & \textcolor{black!55}{\underline{60.7}} & \textcolor{black!55}{\underline{69.1}} & \textcolor{black!55}{60.2} & \textcolor{black!55}{59.2}\\
\cmidrule(lr){1-11}
\hspace{4pt}LensWalk & 53.6 & 51.5 & 67.4 & 35.0 & 62.2 & 58.1 & 51.2 & 67.1 & 50.0 & 57.1\\
\hspace{4pt}Video-RSI & \underline{64.5} & \underline{69.1} & \textbf{78.3} & 40.0 & \underline{71.6} & \underline{64.1} & 54.8 & 65.8 & \underline{68.2} & \textbf{67.3}\\
\specialrule{\lightrulewidth}{0pt}{0pt}
\rowcolor{mvagentrow}
\hspace{4pt}\textbf{MVAgent} & \textbf{69.0} & \textbf{75.0} & 71.7 & \textbf{55.0} & \textbf{73.0} & \textbf{70.3} & \textbf{63.1} & \textbf{73.2} & \textbf{77.3} & \underline{61.2}\\
\noalign{\hbox{\tikz[remember picture,overlay]{\coordinate (detail-association-g2-bottom);}}}
\midrule
\rowcolor{black!8}
\multicolumn{11}{l}{\textit{Qwen3.5-27B}}\\
\noalign{\hbox{\tikz[remember picture,overlay]{\coordinate (detail-association-g3-top);}}}
\hspace{4pt}\textcolor{black!55}{\textit{Single-pass}} & \textcolor{black!55}{\underline{68.5}} & \textcolor{black!55}{\underline{75.0}} & \textcolor{black!55}{\textbf{82.6}} & \textcolor{black!55}{\textbf{48.3}} & \textcolor{black!55}{70.3} & \textcolor{black!55}{71.1} & \textcolor{black!55}{\underline{63.1}} & \textcolor{black!55}{70.5} & \textcolor{black!55}{\textbf{75.0}} & \textcolor{black!55}{\textbf{79.6}}\\
\cmidrule(lr){1-11}
\hspace{4pt}LensWalk & 61.3 & 67.6 & \underline{76.1} & 41.7 & 62.2 & 61.6 & 52.4 & 69.1 & 54.5 & \underline{67.3}\\
\hspace{4pt}Video-RSI & 67.7 & 73.5 & 69.6 & \underline{46.7} & \textbf{78.4} & \underline{71.4} & \textbf{64.3} & \underline{75.2} & \underline{73.9} & \underline{67.3}\\
\specialrule{\lightrulewidth}{0pt}{0pt}
\rowcolor{mvagentrow}
\hspace{4pt}\textbf{MVAgent} & \textbf{69.4} & \textbf{79.4} & 73.9 & \textbf{48.3} & \underline{74.3} & \textbf{71.6} & \textbf{64.3} & \textbf{76.5} & 72.7 & \underline{67.3}\\
\noalign{\hbox{\tikz[remember picture,overlay]{\coordinate (detail-association-g3-bottom);}}}
\midrule
\rowcolor{black!8}
\multicolumn{11}{l}{\textit{GPT-6-sol}}\\
\noalign{\hbox{\tikz[remember picture,overlay]{\coordinate (detail-association-g4-top);}}}
\hspace{4pt}\textcolor{black!55}{\textit{Single-pass}} & \textcolor{black!55}{\underline{75.4}} & \textcolor{black!55}{\underline{79.4}} & \textcolor{black!55}{\underline{87.0}} & \textcolor{black!55}{\underline{50.0}} & \textcolor{black!55}{\underline{85.1}} & \textcolor{black!55}{\underline{76.8}} & \textcolor{black!55}{\underline{69.0}} & \textcolor{black!55}{\textbf{75.2}} & \textcolor{black!55}{\underline{83.0}} & \textcolor{black!55}{\underline{83.7}}\\
\specialrule{\lightrulewidth}{0pt}{0pt}
\rowcolor{mvagentrow}
\hspace{4pt}\textbf{MVAgent} & \textbf{82.7} & \textbf{85.3} & \textbf{95.7} & \textbf{60.0} & \textbf{90.5} & \textbf{77.8} & \textbf{73.8} & \underline{71.8} & \textbf{86.4} & \textbf{87.8}\\
\noalign{\hbox{\tikz[remember picture,overlay]{\coordinate (detail-association-g4-bottom);}}}
\bottomrule
\end{tabularx}
\tikz[remember picture,overlay]{
\draw[black,line width=1.2pt] (detail-association-v1 |- detail-association-top) -- (detail-association-v1 |- detail-association-header-bottom);
\draw[black,line width=1.2pt] (detail-association-v1 |- detail-association-g0-top) -- (detail-association-v1 |- detail-association-g0-bottom);
\draw[black,line width=1.2pt] (detail-association-v1 |- detail-association-g1-top) -- (detail-association-v1 |- detail-association-g1-bottom);
\draw[black,line width=1.2pt] (detail-association-v1 |- detail-association-g2-top) -- (detail-association-v1 |- detail-association-g2-bottom);
\draw[black,line width=1.2pt] (detail-association-v1 |- detail-association-g3-top) -- (detail-association-v1 |- detail-association-g3-bottom);
\draw[black,line width=1.2pt] (detail-association-v1 |- detail-association-g4-top) -- (detail-association-v1 |- detail-association-g4-bottom);
\draw[black,line width=.65pt] (detail-association-v2 |- detail-association-subtop) -- (detail-association-v2 |- detail-association-header-bottom);
\draw[black,line width=.65pt] (detail-association-v2 |- detail-association-g0-top) -- (detail-association-v2 |- detail-association-g0-bottom);
\draw[black,line width=.65pt] (detail-association-v2 |- detail-association-g1-top) -- (detail-association-v2 |- detail-association-g1-bottom);
\draw[black,line width=.65pt] (detail-association-v2 |- detail-association-g2-top) -- (detail-association-v2 |- detail-association-g2-bottom);
\draw[black,line width=.65pt] (detail-association-v2 |- detail-association-g3-top) -- (detail-association-v2 |- detail-association-g3-bottom);
\draw[black,line width=.65pt] (detail-association-v2 |- detail-association-g4-top) -- (detail-association-v2 |- detail-association-g4-bottom);
\draw[black,line width=1.2pt] (detail-association-v6 |- detail-association-top) -- (detail-association-v6 |- detail-association-header-bottom);
\draw[black,line width=1.2pt] (detail-association-v6 |- detail-association-g0-top) -- (detail-association-v6 |- detail-association-g0-bottom);
\draw[black,line width=1.2pt] (detail-association-v6 |- detail-association-g1-top) -- (detail-association-v6 |- detail-association-g1-bottom);
\draw[black,line width=1.2pt] (detail-association-v6 |- detail-association-g2-top) -- (detail-association-v6 |- detail-association-g2-bottom);
\draw[black,line width=1.2pt] (detail-association-v6 |- detail-association-g3-top) -- (detail-association-v6 |- detail-association-g3-bottom);
\draw[black,line width=1.2pt] (detail-association-v6 |- detail-association-g4-top) -- (detail-association-v6 |- detail-association-g4-bottom);
\draw[black,line width=.65pt] (detail-association-v7 |- detail-association-subtop) -- (detail-association-v7 |- detail-association-header-bottom);
\draw[black,line width=.65pt] (detail-association-v7 |- detail-association-g0-top) -- (detail-association-v7 |- detail-association-g0-bottom);
\draw[black,line width=.65pt] (detail-association-v7 |- detail-association-g1-top) -- (detail-association-v7 |- detail-association-g1-bottom);
\draw[black,line width=.65pt] (detail-association-v7 |- detail-association-g2-top) -- (detail-association-v7 |- detail-association-g2-bottom);
\draw[black,line width=.65pt] (detail-association-v7 |- detail-association-g3-top) -- (detail-association-v7 |- detail-association-g3-bottom);
\draw[black,line width=.65pt] (detail-association-v7 |- detail-association-g4-top) -- (detail-association-v7 |- detail-association-g4-bottom);
}

\end{table*}

\begin{table*}[tp]
\centering
\caption{\textbf{Detailed results on CrossVid~\cite{li2026crossvid}.} Tasks: behavioral understanding (BU), narrative comprehension (NC), culinary comparison (CC), procedural error analysis (PEA), plot inference (PI), functional step alignment (FSA), procedural step sequencing (PSS), multi-view spatial reasoning (MSR), multi-view object counting (MOC), comparative culinary QA (CCQA). C.Avg, T.Avg, and M.Avg denote mean scores for comparative analysis, temporal understanding, and multi-view reasoning, respectively.}
\label{app:crossvid-detailed}
\footnotesize
\setlength{\tabcolsep}{1.1pt}
\renewcommand{\arraystretch}{1.24}
\setlength{\aboverulesep}{1pt}
\setlength{\belowrulesep}{1pt}
\setlength{\cmidrulesep}{1pt}
\begin{tabularx}{\textwidth}{>{\raggedright\arraybackslash}p{68pt}!{\tikz[remember picture,overlay]{\coordinate (detail-cross-v1);}}>{\centering\arraybackslash}X!{\tikz[remember picture,overlay]{\coordinate (detail-cross-v2);}}>{\centering\arraybackslash}X>{\centering\arraybackslash}X>{\centering\arraybackslash}X>{\centering\arraybackslash}X!{\tikz[remember picture,overlay]{\coordinate (detail-cross-v6);}}>{\centering\arraybackslash}X!{\tikz[remember picture,overlay]{\coordinate (detail-cross-v7);}}>{\centering\arraybackslash}X>{\centering\arraybackslash}X>{\centering\arraybackslash}X!{\tikz[remember picture,overlay]{\coordinate (detail-cross-v10);}}>{\centering\arraybackslash}X!{\tikz[remember picture,overlay]{\coordinate (detail-cross-v11);}}>{\centering\arraybackslash}X>{\centering\arraybackslash}X!{\tikz[remember picture,overlay]{\coordinate (detail-cross-v13);}}>{\centering\arraybackslash}p{34pt}}
\toprule
\noalign{\hbox{\tikz[remember picture,overlay]{\coordinate (detail-cross-top);}}}
\multirow{2}{=}[-0.7ex]{\centering\textbf{Method}} & \multicolumn{5}{c}{\textbf{Comparative Analysis}} & \multicolumn{4}{c}{\textbf{Temporal Understanding}} & \multicolumn{3}{c}{\textbf{Multi-view Reasoning}} & \multicolumn{1}{c}{\textbf{Free-form}}\\
\cmidrule{2-14}
\noalign{\hbox{\tikz[remember picture,overlay]{\coordinate (detail-cross-subtop);}}}
 & \textbf{C.Avg} & \textbf{BU} & \textbf{NC} & \textbf{CC} & \textbf{PEA} & \textbf{T.Avg} & \textbf{PI} & \textbf{FSA} & \textbf{PSS} & \textbf{M.Avg} & \textbf{MSR} & \textbf{MOC} & \textbf{CCQA}\\
\noalign{\hbox{\tikz[remember picture,overlay]{\coordinate (detail-cross-header-bottom);}}}
\midrule
\rowcolor{black!8}
\multicolumn{14}{l}{\textit{AgentCVR-8B}~\cite{qiu2026agentcvr}}\\
\noalign{\hbox{\tikz[remember picture,overlay]{\coordinate (detail-cross-g0-top);}}}
\hspace{4pt}AgentCVR & 48.4 & 35.3 & 45.8 & 80.3 & 32.2 & 22.1 & 28.0 & 22.3 & 15.9 & 36.8 & 36.2 & 37.4 & 54.8\\
\noalign{\hbox{\tikz[remember picture,overlay]{\coordinate (detail-cross-g0-bottom);}}}
\midrule
\rowcolor{black!8}
\multicolumn{14}{l}{\textit{Qwen3.5-9B}~\cite{qwenteam2026qwen35}}\\
\noalign{\hbox{\tikz[remember picture,overlay]{\coordinate (detail-cross-g1-top);}}}
\hspace{4pt}\textcolor{black!55}{\textit{Single-pass}} & \textcolor{black!55}{30.0} & \textcolor{black!55}{16.4} & \textcolor{black!55}{24.6} & \textcolor{black!55}{\underline{52.3}} & \textcolor{black!55}{26.8} & \textcolor{black!55}{\underline{28.0}} & \textcolor{black!55}{\textbf{73.0}} & \textcolor{black!55}{4.3} & \textcolor{black!55}{6.6} & \textcolor{black!55}{31.5} & \textcolor{black!55}{30.1} & \textcolor{black!55}{\underline{32.9}} & \textcolor{black!55}{\textbf{50.1}}\\
\cmidrule(lr){1-14}
\hspace{4pt}LensWalk~\cite{li2026lenswalk} & 29.8 & 15.9 & \underline{35.0} & 44.5 & 23.8 & 25.8 & 38.0 & \textbf{25.1} & \underline{14.3} & 13.9 & 11.7 & 16.1 & \underline{39.9}\\
\hspace{4pt}Video-RSI~\cite{luo2026videorsi} & \underline{34.4} & \underline{27.2} & 32.9 & 35.3 & \textbf{42.1} & 25.0 & \underline{63.0} & 4.2 & 7.7 & \underline{32.4} & \underline{30.7} & \textbf{34.2} & 30.3\\
\specialrule{\lightrulewidth}{0pt}{0pt}
\rowcolor{mvagentrow}
\hspace{4pt}\textbf{MVAgent} & \textbf{47.0} & \textbf{40.9} & \textbf{53.3} & \textbf{61.5} & \underline{32.2} & \textbf{33.6} & 57.0 & \underline{15.3} & \textbf{28.6} & \textbf{33.9} & \textbf{35.0} & 32.9 & 28.6\\
\noalign{\hbox{\tikz[remember picture,overlay]{\coordinate (detail-cross-g1-bottom);}}}
\midrule
\rowcolor{black!8}
\multicolumn{14}{l}{\textit{Qwen3.5-35B-A3B}}\\
\noalign{\hbox{\tikz[remember picture,overlay]{\coordinate (detail-cross-g2-top);}}}
\hspace{4pt}\textcolor{black!55}{\textit{Single-pass}} & \textcolor{black!55}{\underline{39.9}} & \textcolor{black!55}{25.9} & \textcolor{black!55}{\underline{39.5}} & \textcolor{black!55}{\textbf{66.1}} & \textcolor{black!55}{28.0} & \textcolor{black!55}{28.5} & \textcolor{black!55}{\textbf{79.0}} & \textcolor{black!55}{4.3} & \textcolor{black!55}{2.2} & \textcolor{black!55}{19.9} & \textcolor{black!55}{16.6} & \textcolor{black!55}{23.2} & \textcolor{black!55}{\textbf{47.6}}\\
\cmidrule(lr){1-14}
\hspace{4pt}LensWalk & 23.4 & 28.4 & 23.7 & 23.4 & 18.0 & 25.2 & 45.0 & \underline{17.5} & \underline{13.2} & 10.0 & 13.5 & 6.5 & 33.2\\
\hspace{4pt}Video-RSI & 38.8 & \underline{34.1} & 38.3 & 38.1 & \textbf{44.8} & \underline{32.1} & \underline{73.0} & 10.8 & 12.6 & \underline{33.4} & \underline{30.7} & \underline{36.1} & 27.3\\
\specialrule{\lightrulewidth}{0pt}{0pt}
\rowcolor{mvagentrow}
\hspace{4pt}\textbf{MVAgent} & \textbf{46.0} & \textbf{40.9} & \textbf{52.4} & \underline{56.9} & \underline{33.7} & \textbf{40.5} & 62.0 & \textbf{19.3} & \textbf{40.1} & \textbf{38.3} & \textbf{39.3} & \textbf{37.4} & \underline{44.5}\\
\noalign{\hbox{\tikz[remember picture,overlay]{\coordinate (detail-cross-g2-bottom);}}}
\midrule
\rowcolor{black!8}
\multicolumn{14}{l}{\textit{Qwen3.5-27B}}\\
\noalign{\hbox{\tikz[remember picture,overlay]{\coordinate (detail-cross-g3-top);}}}
\hspace{4pt}\textcolor{black!55}{\textit{Single-pass}} & \textcolor{black!55}{\textbf{51.3}} & \textcolor{black!55}{41.4} & \textcolor{black!55}{43.1} & \textcolor{black!55}{\textbf{73.4}} & \textcolor{black!55}{\textbf{47.5}} & \textcolor{black!55}{31.1} & \textcolor{black!55}{\textbf{83.0}} & \textcolor{black!55}{4.9} & \textcolor{black!55}{5.5} & \textcolor{black!55}{25.6} & \textcolor{black!55}{20.9} & \textcolor{black!55}{\underline{30.3}} & \textcolor{black!55}{\textbf{66.6}}\\
\cmidrule(lr){1-14}
\hspace{4pt}LensWalk & 36.2 & 31.5 & \underline{48.5} & 49.5 & 15.3 & 36.9 & 67.0 & \textbf{32.9} & 11.0 & 17.6 & 16.6 & 18.7 & \underline{52.2}\\
\hspace{4pt}Video-RSI & \underline{50.1} & \textbf{49.1} & 47.6 & 59.6 & \underline{44.1} & \underline{45.9} & 75.0 & \underline{31.9} & \underline{30.8} & \underline{31.4} & \underline{34.4} & 28.4 & 44.4\\
\specialrule{\lightrulewidth}{0pt}{0pt}
\rowcolor{mvagentrow}
\hspace{4pt}\textbf{MVAgent} & 50.0 & \underline{46.1} & \textbf{52.4} & \underline{65.6} & 36.0 & \textbf{53.1} & \underline{80.0} & 26.6 & \textbf{52.7} & \textbf{42.4} & \textbf{43.6} & \textbf{41.3} & 48.4\\
\noalign{\hbox{\tikz[remember picture,overlay]{\coordinate (detail-cross-g3-bottom);}}}
\midrule
\rowcolor{black!8}
\multicolumn{14}{l}{\textit{GPT-6-sol}~\cite{openai2026gpt6sol}}\\
\noalign{\hbox{\tikz[remember picture,overlay]{\coordinate (detail-cross-g4-top);}}}
\hspace{4pt}\textcolor{black!55}{\textit{Single-pass}} & \textcolor{black!55}{\underline{65.1}} & \textcolor{black!55}{\textbf{66.8}} & \textcolor{black!55}{\underline{57.8}} & \textcolor{black!55}{\underline{82.6}} & \textcolor{black!55}{\textbf{53.3}} & \textcolor{black!55}{\underline{53.1}} & \textcolor{black!55}{\underline{87.0}} & \textcolor{black!55}{\underline{11.2}} & \textcolor{black!55}{\underline{61.0}} & \textcolor{black!55}{\underline{43.8}} & \textcolor{black!55}{\textbf{54.0}} & \textcolor{black!55}{\underline{33.5}} & \textcolor{black!55}{\textbf{65.4}}\\
\specialrule{\lightrulewidth}{0pt}{0pt}
\rowcolor{mvagentrow}
\hspace{4pt}\textbf{MVAgent} & \textbf{67.6} & \underline{55.6} & \textbf{68.9} & \textbf{92.7} & \textbf{53.3} & \textbf{68.2} & \textbf{100.0} & \textbf{21.1} & \textbf{83.5} & \textbf{51.6} & \underline{51.5} & \textbf{51.6} & \underline{60.8}\\
\noalign{\hbox{\tikz[remember picture,overlay]{\coordinate (detail-cross-g4-bottom);}}}
\bottomrule
\end{tabularx}
\tikz[remember picture,overlay]{
\draw[black,line width=1.2pt] (detail-cross-v1 |- detail-cross-top) -- (detail-cross-v1 |- detail-cross-header-bottom);
\draw[black,line width=1.2pt] (detail-cross-v1 |- detail-cross-g0-top) -- (detail-cross-v1 |- detail-cross-g0-bottom);
\draw[black,line width=1.2pt] (detail-cross-v1 |- detail-cross-g1-top) -- (detail-cross-v1 |- detail-cross-g1-bottom);
\draw[black,line width=1.2pt] (detail-cross-v1 |- detail-cross-g2-top) -- (detail-cross-v1 |- detail-cross-g2-bottom);
\draw[black,line width=1.2pt] (detail-cross-v1 |- detail-cross-g3-top) -- (detail-cross-v1 |- detail-cross-g3-bottom);
\draw[black,line width=1.2pt] (detail-cross-v1 |- detail-cross-g4-top) -- (detail-cross-v1 |- detail-cross-g4-bottom);
\draw[black,line width=1.2pt] (detail-cross-v6 |- detail-cross-top) -- (detail-cross-v6 |- detail-cross-header-bottom);
\draw[black,line width=1.2pt] (detail-cross-v6 |- detail-cross-g0-top) -- (detail-cross-v6 |- detail-cross-g0-bottom);
\draw[black,line width=1.2pt] (detail-cross-v6 |- detail-cross-g1-top) -- (detail-cross-v6 |- detail-cross-g1-bottom);
\draw[black,line width=1.2pt] (detail-cross-v6 |- detail-cross-g2-top) -- (detail-cross-v6 |- detail-cross-g2-bottom);
\draw[black,line width=1.2pt] (detail-cross-v6 |- detail-cross-g3-top) -- (detail-cross-v6 |- detail-cross-g3-bottom);
\draw[black,line width=1.2pt] (detail-cross-v6 |- detail-cross-g4-top) -- (detail-cross-v6 |- detail-cross-g4-bottom);
\draw[black,line width=1.2pt] (detail-cross-v10 |- detail-cross-top) -- (detail-cross-v10 |- detail-cross-header-bottom);
\draw[black,line width=1.2pt] (detail-cross-v10 |- detail-cross-g0-top) -- (detail-cross-v10 |- detail-cross-g0-bottom);
\draw[black,line width=1.2pt] (detail-cross-v10 |- detail-cross-g1-top) -- (detail-cross-v10 |- detail-cross-g1-bottom);
\draw[black,line width=1.2pt] (detail-cross-v10 |- detail-cross-g2-top) -- (detail-cross-v10 |- detail-cross-g2-bottom);
\draw[black,line width=1.2pt] (detail-cross-v10 |- detail-cross-g3-top) -- (detail-cross-v10 |- detail-cross-g3-bottom);
\draw[black,line width=1.2pt] (detail-cross-v10 |- detail-cross-g4-top) -- (detail-cross-v10 |- detail-cross-g4-bottom);
\draw[black,line width=1.2pt] (detail-cross-v13 |- detail-cross-top) -- (detail-cross-v13 |- detail-cross-header-bottom);
\draw[black,line width=1.2pt] (detail-cross-v13 |- detail-cross-g0-top) -- (detail-cross-v13 |- detail-cross-g0-bottom);
\draw[black,line width=1.2pt] (detail-cross-v13 |- detail-cross-g1-top) -- (detail-cross-v13 |- detail-cross-g1-bottom);
\draw[black,line width=1.2pt] (detail-cross-v13 |- detail-cross-g2-top) -- (detail-cross-v13 |- detail-cross-g2-bottom);
\draw[black,line width=1.2pt] (detail-cross-v13 |- detail-cross-g3-top) -- (detail-cross-v13 |- detail-cross-g3-bottom);
\draw[black,line width=1.2pt] (detail-cross-v13 |- detail-cross-g4-top) -- (detail-cross-v13 |- detail-cross-g4-bottom);
\draw[black,line width=.65pt] (detail-cross-v2 |- detail-cross-subtop) -- (detail-cross-v2 |- detail-cross-header-bottom);
\draw[black,line width=.65pt] (detail-cross-v2 |- detail-cross-g0-top) -- (detail-cross-v2 |- detail-cross-g0-bottom);
\draw[black,line width=.65pt] (detail-cross-v2 |- detail-cross-g1-top) -- (detail-cross-v2 |- detail-cross-g1-bottom);
\draw[black,line width=.65pt] (detail-cross-v2 |- detail-cross-g2-top) -- (detail-cross-v2 |- detail-cross-g2-bottom);
\draw[black,line width=.65pt] (detail-cross-v2 |- detail-cross-g3-top) -- (detail-cross-v2 |- detail-cross-g3-bottom);
\draw[black,line width=.65pt] (detail-cross-v2 |- detail-cross-g4-top) -- (detail-cross-v2 |- detail-cross-g4-bottom);
\draw[black,line width=.65pt] (detail-cross-v7 |- detail-cross-subtop) -- (detail-cross-v7 |- detail-cross-header-bottom);
\draw[black,line width=.65pt] (detail-cross-v7 |- detail-cross-g0-top) -- (detail-cross-v7 |- detail-cross-g0-bottom);
\draw[black,line width=.65pt] (detail-cross-v7 |- detail-cross-g1-top) -- (detail-cross-v7 |- detail-cross-g1-bottom);
\draw[black,line width=.65pt] (detail-cross-v7 |- detail-cross-g2-top) -- (detail-cross-v7 |- detail-cross-g2-bottom);
\draw[black,line width=.65pt] (detail-cross-v7 |- detail-cross-g3-top) -- (detail-cross-v7 |- detail-cross-g3-bottom);
\draw[black,line width=.65pt] (detail-cross-v7 |- detail-cross-g4-top) -- (detail-cross-v7 |- detail-cross-g4-bottom);
\draw[black,line width=.65pt] (detail-cross-v11 |- detail-cross-subtop) -- (detail-cross-v11 |- detail-cross-header-bottom);
\draw[black,line width=.65pt] (detail-cross-v11 |- detail-cross-g0-top) -- (detail-cross-v11 |- detail-cross-g0-bottom);
\draw[black,line width=.65pt] (detail-cross-v11 |- detail-cross-g1-top) -- (detail-cross-v11 |- detail-cross-g1-bottom);
\draw[black,line width=.65pt] (detail-cross-v11 |- detail-cross-g2-top) -- (detail-cross-v11 |- detail-cross-g2-bottom);
\draw[black,line width=.65pt] (detail-cross-v11 |- detail-cross-g3-top) -- (detail-cross-v11 |- detail-cross-g3-bottom);
\draw[black,line width=.65pt] (detail-cross-v11 |- detail-cross-g4-top) -- (detail-cross-v11 |- detail-cross-g4-bottom);
}
\end{table*}

Due to space constraints, the main paper omits detailed task-level scores for CVBench and CrossVid; we provide them here.

\paragraph{CVBench.}
Table~\ref{app:cvbench-detailed} follows CVBench's original result-table groupings~\cite{zhu2026cvbench}. The reasoning, object association, and event association averages correspond to Reas, Obj, and Evt in the main table, respectively.

\paragraph{CrossVid.}
Table~\ref{app:crossvid-detailed} follows the task and dimension layout of the original CrossVid results table~\cite{li2026crossvid}. CCQA uses the same saved DeepSeek-V4.1-Flash~\cite{deepseek2026flashdocs} judgments as the main table.

\subsection{Additional Analysis}
\label{app:additional-analysis}
\begin{figure}[!tbp]
\centering
\includegraphics[width=\columnwidth]{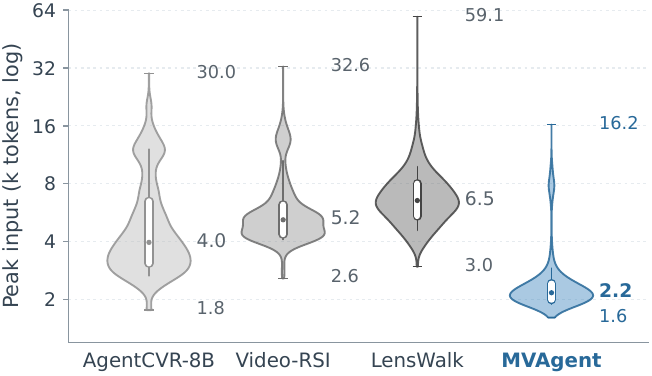}
\caption{\textbf{Peak planning input per question.} Values label the maximum, median, and minimum (k tokens). Dots, white bars, and whiskers mark the median, 25th--75th, and 10th--90th percentiles. Violins show relative density in log space and extend to the observed extrema.}
\label{app:planning-input-distribution}
\end{figure}

\paragraph{Planning-input distributions.}
Figure~\ref{app:planning-input-distribution} shows the maximum native input-token count across each question's planning calls, including GlobalAgent and all LocalAgents for MVAgent. Qwen3.5~\cite{qwenteam2026qwen35} results pool the three model sizes; AgentCVR~\cite{qiu2026agentcvr} uses its released 8B configuration. The median peak is 2.2k tokens for MVAgent and 4.0--6.5k for the other systems.

\section{System Messages and Prompt Templates}
\label{app:prompts}
We reproduce the current implementation's fixed prompt components, with agent and field names aligned to the main paper. Braced placeholders are filled at runtime. Source formatting and JSON limit annotations are preserved; long lines wrap for display. Historical runs may use earlier templates.

\subsection{Prompt assembly and availability}
Each decision combines a system message, a user template, and the currently available action descriptions. The question, ordered source metadata, and global or local history fill the placeholders.

\subsection{GlobalAgent decisions and actions}

\begin{promptbox}{GlobalAgent: system message}{prompts/global_system.tex}
\begin{Verbatim}
You are a coordinator for multi-video question answering. You cannot view videos directly. Your visual evidence consists only of successful single-video summaries and completed joint-watch results in History. Preserve video identities and choose one available action that either obtains missing evidence or returns the final answer.
\end{Verbatim}

\end{promptbox}

\begin{promptbox}{GlobalAgent: decision user template}{prompts/global_decision.tex}
\begin{Verbatim}
# Multi-video evidence decision

## Task
Coordinate the complete question across videos and choose one available action that either obtains missing evidence or returns the answer.

## Inputs

### Question
{question}

### Videos
{videos}

## Available actions
{action_prompt_blocks}

## History
{action_history}

## Rules
- Single-video agents' Observe actions and intermediate visual results are private. Use only facts stated in public summaries and completed joint-watch results.
- Return one decision with a concise evidence-based `reason`, one available `action`, and exactly that action's `parameters`.
- Treat `reason` as a brief decision basis, not an open-ended reasoning transcript.
- Treat the IDs in Videos as an exact allow-list. Copy each `video_id` exactly; do not use an unlisted ID or a positional or semantic alias such as `first`, `middle`, or `reference`.
- Within one `analyze_videos` or `watch_videos` action, each listed video ID may appear at most once.
- Use the durations in Videos when choosing source ranges.

## Decision strategy
None

## Output contract
Choose the next action and return its JSON object only.
\end{Verbatim}

\end{promptbox}

\begin{promptbox}{GlobalAgent action: \analyze{}}{prompts/global_analyze_action.tex}
\begin{Verbatim}
### Action: analyze_videos

Function:
Ask one or more selected single-video agents to analyze their own videos separately. Each analysis is executed only on its selected video and must be independently understandable from its instruction.

Fields:
- `reason`: Give a brief basis for using separate per-video analysis.
- `parameters.localagent_request`: Return one or more objects containing one existing `video_id` and one non-empty local `instruction`.

Rules:
- Each `video_id` must be copied exactly from the IDs listed under Videos. Never use an unlisted alias or position label, and never request the same video ID more than once in one `analyze_videos` action.
- The selected single-video agents run separately. They never watch multiple videos together and cannot perform joint cross-video visual comparison.
- Each single-video agent sees only its own video and local instruction; it does not receive the complete question, Global History, other videos, or sibling analyses and results.
- Every local instruction must be self-contained and answerable from the selected video alone.
- Every requested output must be visually observable in the selected video. Ask for underlying visible mechanics or attributes, not an external category-to-score table, a standard difficulty value that is not displayed, a cross-video mapping, or a computed answer.
- Across this action, reserve at most {analyze_instruction_budget} characters for all instructions. For N requested videos, each instruction must also stay within floor({analyze_instruction_budget} / N) characters.
- Analyses emitted in one decision are knowledge-independent: no single-video agent can see another analysis or use a sibling result from the same batch.

Example:
{
  "reason": "<brief basis for using separate video analysis; nonblank; at most 1200 characters>",
  "action": "analyze_videos",
  "parameters": {
    "localagent_request": [
      {
        "video_id": "<one existing video ID>",
        "instruction": "<one self-contained, visually answerable instruction for that video; nonblank; at most 1600 characters>"
      }
    ]
  }
}
\end{Verbatim}

\end{promptbox}

\begin{promptbox}{GlobalAgent action: \watch{}}{prompts/global_watch_action.tex}
\begin{Verbatim}
### Action: watch_videos

Function:
Select clips from multiple (>1) input videos and observe them jointly when detailed visual information must be compared.

Fields:
- `reason`: State why the required cross-video visual relationship must be watched jointly rather than analyzed separately.
- `parameters.instruction`: Give one self-contained visual-analysis instruction. This exact string is the only text prompt sent to the visual model, so name the selected videos by ID, state exactly which visible relationship, difference, correspondence, phase, motion, appearance, or spatial detail to analyze, and request the desired concise natural-language analysis directly.
- `parameters.videos`: Return at least two video selections. Each selection contains one input-owned `video_id` and exactly one `clip` in `[start_sec, end_sec]` form.

Rules:
- Select more than one distinct input video, but no more than the number listed under Videos or {max_videos_per_watch}, whichever is smaller.
- Every `video_id` must be copied exactly from the IDs listed under Videos; never invent, rename, or infer a video ID.
- Every selected clip is supplied to the same visual-model call. Select each video ID at most once and never return more than one clip for a video.
- Each clip must provide complete start and end seconds inside its source video's duration.
- The number of selected videos must not exceed the visual model's limit of {max_video_inputs_per_request} video inputs.
- Runtime chooses sampling automatically; never include `fps` in this action's parameters. With fixed visual input capacity, longer total selected duration can reveal more of the overall sequence, but makes observation sparser and increases the risk of missing brief or fine-grained details.
- The visual model returns one unstructured natural-language string as evidence for the next multi-video decision. Do not ask it to choose the final answer.

Example:
{
  "reason": "<brief basis for joint video analysis; nonblank; at most 1200 characters>",
  "action": "watch_videos",
  "parameters": {
    "instruction": "<one self-contained visual-analysis instruction naming the selected video IDs and the exact relationship to compare; nonblank; at most 1600 characters>",
    "videos": [
      {
        "video_id": "<one selected video ID>",
        "clip": ["<start_sec>", "<end_sec>"]
      },
      {
        "video_id": "<another selected video ID>",
        "clip": ["<start_sec>", "<end_sec>"]
      }
    ]
  }
}
\end{Verbatim}

\end{promptbox}

\begin{promptbox}{GlobalAgent action: \answer{}}{prompts/global_answer_action.tex}
\begin{Verbatim}
### Action: answer

Function:
Return the final answer directly from the question and accumulated History.

Fields:
- `reason`: Give a concise evidence-based explanation, including any material uncertainty.
- `parameters.answer`: The requested answer representation, as a non-empty string.

Rules:
- Choose answer when the judgment is supported or further checks cannot add useful information.
- Once answering, submit the best-supported judgment; do not replace the requested result with a discussion of missing evidence.
- Infer the answer representation from the complete question. When options are present, use only the selected option label or explicitly requested option text in `parameters.answer`.

Example:
{
  "reason": "<brief basis for this answer; nonblank>",
  "action": "answer",
  "parameters": {"answer": "<the requested answer representation; nonblank; at most 2400 characters>"}
}
\end{Verbatim}

\end{promptbox}

At the global round limit, the strategy field receives the following instruction (Sec.~\ref{app:validation-fallback}).
\begin{promptbox}{GlobalAgent: terminal strategy insertion}{prompts/global_terminal_strategy.tex}
\begin{Verbatim}
No further evidence collection is available. Submit the best-supported final answer now.
\end{Verbatim}

\end{promptbox}

\subsection{LocalAgent decisions and reporting}

\begin{promptbox}{LocalAgent: system message}{prompts/local_system.tex}
\begin{Verbatim}
You are the text-only Planner for one single-video evidence agent. You cannot view the video directly. Same-video Memory is your only source of visible facts. Choose one available action that either inspects another source-video range or reports retained evidence for the active instruction.
\end{Verbatim}

\end{promptbox}

\begin{promptbox}{LocalAgent: decision user template}{prompts/local_decision.tex}
\begin{Verbatim}
# Single-video evidence decision

## Task
Choose one available action for the active instruction and accumulated same-video evidence.

## Inputs

### Instruction
{instruction}

### Video context
- Duration: {duration_sec} seconds

## Memory
{memory_view}

## Available actions
{action_prompt_blocks}

## Rules
- Return one available `action` with exactly its matching `parameters`.
- Obey the selected action's field contract.

## Decision strategy
None

## Output contract
Choose whether to gather more evidence or finish, and return the action JSON object only.
\end{Verbatim}

\end{promptbox}

\begin{promptbox}{LocalAgent action: \observe{}}{prompts/local_observe_action.tex}
\begin{Verbatim}
### Action: observe

Function:
Semantically interpret one or more source-video ranges for the local instruction. Runtime records one semantic observation per requested range (exact source range, actual FPS, objective text, optional uncertainty); the next Planner call sees them as compact natural-language Memory lines, never as raw dictionaries.

Fields:
- `parameters.what` (string): One non-empty, visually answerable question the visual model must resolve.
- `parameters.where` (array of ranges): One to {max_ranges_per_action} ordered, non-overlapping `[start, end]` source-second ranges inside the video duration.
- `parameters.fps` (number): One planned sampling rate in [{min_observe_fps}, {max_observe_fps}] FPS shared by every requested range, using at most one decimal place (0.1 increments).

Rules:
- One observe range may cover any portion of the video up to the full duration; there is no fixed duration cap.
- The visual model allows at most {max_video_frames_per_request} input frames per call. Runtime adjusts effective sampling when needed to stay within that capacity and provide a valid temporal video input while preserving the requested source boundaries.
- Runtime validates source ranges.

Example:
{
  "action": "observe",
  "parameters": {
    "what": "<one visually answerable question the visual model must resolve; nonblank; at most 1600 characters>",
    "where": [["<start_sec>", "<end_sec>"]],
    "fps": "<planned sampling FPS in [0.1, 25.0], using 0.1 increments>"
  }
}
\end{Verbatim}

\end{promptbox}

\begin{promptbox}{LocalAgent action: \finish{}}{prompts/local_finish_action.tex}
\begin{Verbatim}
### Action: finish

Function:
Stop this request and report answer-relevant observed evidence to the multi-video coordinator as one summary carried by this action's parameters. Finish is available only after Runtime has retained at least one same-video observation, including observations from earlier requests. The multi-video coordinator does not see this single-video agent's internal Observe actions, ranges, sampling rates, or separate observation results; finish itself never enters Memory.

Fields:
- `parameters.summary` (string): One non-empty, self-contained summary grounded only in retained same-video observations. Include any answer-relevant uncertainty in this same string. Do not expose internal Observe calls, source ranges, FPS values, planning steps, or tool details. Do not propose further actions, choose answer options, or include planning text.

Rules:
- Return exactly this one field.

Example:
{
  "action": "finish",
  "parameters": {
    "summary": "<non-empty summary grounded in retained same-video observations; nonblank; at most 3200 characters>"
  }
}
\end{Verbatim}

\end{promptbox}

The following template generates the report at the local step limit (Sec.~\ref{app:validation-fallback}).
\begin{promptbox}{LocalAgent: terminal summary user template}{prompts/local_terminal_finish.tex}
\begin{Verbatim}
# Final video summary

## Task
The supplied Memory contains records from earlier observations of one source video. Summarize the retained visible evidence relevant to the active Instruction.

## Inputs

### Instruction
{instruction}

### Memory
{memory_view}

## Evidence rules
- Treat Memory as the only evidence for visible facts. Failed observations and invalid decisions are feedback, not evidence.
- Write one non-empty, self-contained `summary` grounded only in retained same-video observations.
- Include any answer-relevant uncertainty in the same summary.
- Do not expose internal Observe calls, source ranges, FPS values, planning steps, or tool details.

## Output contract
Return one JSON object containing only a non-empty `summary`.

Valid example:
{
  "summary": "<self-contained summary grounded in retained observations; nonblank; at most 3200 characters>"
}
\end{Verbatim}

\end{promptbox}

\subsection{Visual observation and joint inspection}
Each selected range is inspected separately using its source ID, clip, and the following user template, without a system message.
\begin{promptbox}{Visual observation: user template}{prompts/visual_observation.tex}
\begin{Verbatim}
# Clip observation

## Task
The supplied video input represents one selected interval from a source video. Answer the focused Evidence request below using visible content from this clip.

## Clip context
- Full source duration: {duration_sec} seconds
- Supplied source interval: {source_start_sec}–{source_end_sec} seconds
- Clip-local interval: 0.0–{clip_duration_sec} seconds
- Timeline mapping: source_time = {source_start_sec} + clip_time
- Actual sampling: {fps} FPS

The supplied clip already represents the requested source interval; its local timeline starts at 0.0.

## Evidence request
{what}

## Evidence rules
- Answer the evidence request first and ignore details that do not help resolve it.
- When relevant, report visible objects, actions, interactions, scene changes, counts, attributes, spatial relations, on-screen text, and temporal order.
- Distinguish a visibly absent cue from a cue that cannot be determined because of sampling, occlusion, or missing detail.
- Ground the response only in the supplied clip; do not infer events outside it.
- When time localization is requested or useful for a narrower follow-up, report source-video seconds using the mapping above. Otherwise omit timestamps and frame indexes.

## Output contract
Return one JSON object containing only `text` and `uncertainty`.

{
  "text": "<compact visible evidence that addresses the request; nonblank; at most 2000 characters>",
  "uncertainty": "<material ambiguity or limitation; empty string when none; at most 400 characters>"
}
\end{Verbatim}

\end{promptbox}

\paragraph{Joint observation.}
Joint inspection has no fixed template. It receives the ordered clips, source labels \texttt{This is \{video\_id\}}, and GlobalAgent's comparison instruction, which requests visual evidence rather than a final answer.

\subsection{Structured-output prompts}
The repair prompt in Sec.~\ref{app:validation-fallback} receives the serialized original messages, required schema, validation feedback, and raw prior response.
\begin{promptbox}{Structured text output: one repair request}{prompts/format_repair.tex}
\begin{Verbatim}
Repair the previous response so it satisfies the original request and the required JSON Schema. Preserve identifiers and constraints from the original request. Return the smallest complete object: condense free-text fields instead of continuing or repeating the previous analysis. Return JSON only.

Original request:
{original_messages_json}

Required JSON Schema:
{json_schema_json}

Validation error:
{validation_error}

Previous response:
{previous_response}
\end{Verbatim}

\end{promptbox}

Backends using textual schema guidance append this instruction to the first system/developer message, or insert it as a system message if neither exists. Native-schema backends need not add it. The default schema name is \texttt{mvagent\_response}.
\begin{promptbox}{Provider-conditional schema instruction}{prompts/schema_wrapper.tex}
\begin{Verbatim}
Return exactly one JSON object matching the following JSON Schema. Do not add Markdown fences or explanatory text.

Schema name: {schema_name}
JSON Schema:
{json_schema_json}
\end{Verbatim}

\end{promptbox}

\section{Qualitative Cases}
\label{app:cases}
We group the traces by outcome and ability. Teal marks success and amber failure. Boxes quote saved global decisions and selected local steps; [\ldots] marks omissions and whitespace is normalized. IDs, actions, temporal ranges, sampling rates, and answers retain their recorded values. Reports and joint observations are model outputs; ground truth is the benchmark annotation.

\subsection{Successful cases}
\label{app:success-cases}

\subsubsection{Multi-round local observation}
\begin{figure}[!t]
\centering
\includegraphics[width=\columnwidth]{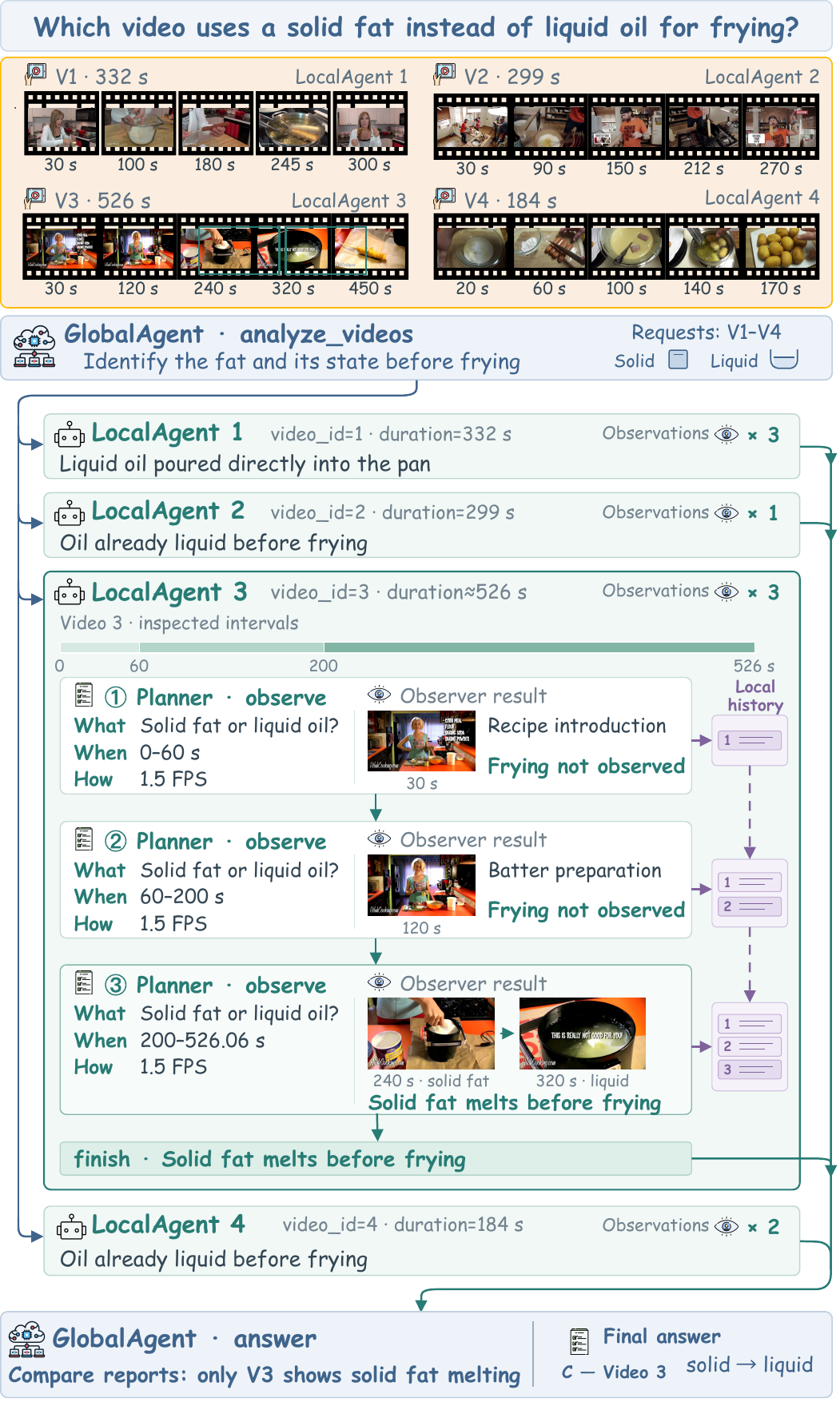}
\caption{\textbf{Multi-round local observation.} LocalAgent 3 locates the relevant evidence after three observations.}
\label{app:local-multiround-case}
\end{figure}

Figure~\ref{app:local-multiround-case} shows three \observe{} actions within one request. LocalAgent retains earlier observations to locate the frying step; GlobalAgent compares the resulting per-video reports.

\subsubsection{Perception and attribute discrimination}
\begin{casebox}{agentteal}{Task: MVU-Eval (OR)}
\textbf{Question:} In the following video, which performer is playing the violin?\\
\textbf{Options:}\par
\casechoices{\textbf{A.} Video 1}{\textbf{B.} Video 2}{\textbf{C.} Video 5}{\textbf{D.} Video 6}
\textbf{Ground truth:} B

\caseturn{Turn 1 / GlobalAgent}{\analyze{}}
\textbf{Reason:} To identify which performer is playing the violin, I need to examine the instruments being played in the candidate videos (1, 2, 5, and 6). [...]\par
\textbf{video\_id:} \texttt{video\_1, video\_2, video\_5, video\_6}\par
\textbf{instruction:} Identify the musical instrument being played by the performer in this video. [...]\par
\textbf{Report / video\_1:} No musical instrument is visible in the video. [...]\par
\textbf{LocalAgent / video\_2:} local steps\par
\textbf{Step 1 / action:} \observe{}\par
\textbf{where:} \texttt{[[0.0, 10.01]]}\par
\textbf{fps:} \texttt{5.2}\par
\textbf{what:} What musical instrument is being played by the performer? [...]\par
\textbf{Observation:} The performer is playing a violin. The instrument has a traditional wooden body with a warm brown finish, four strings, a black fingerboard, and a scroll at the top of [...] with standard violin technique. [...]\par
\textbf{Action:} \finish{}\par
\textbf{Report / video\_2:} The performer is playing a violin. [...]\par
\textbf{Report / video\_5:} The performer is playing a trumpet, a brass instrument with a shiny metallic (likely silver or chrome) finish. [...]\par
\textbf{Report / video\_6:} The performer is playing a clarinet. [...]\par
\caseturn{Turn 2 / GlobalAgent}{\answer{}}
\textbf{Reason:} The analysis of video\_2 explicitly confirms that the performer is playing a violin, describing its appearance and playing technique. [...]\par
\nopagebreak[4]
\casefinal{B}

\end{casebox}

\subsubsection{Quantitative comparison}
\begin{casebox}{agentteal}{Task: MVU-Eval (Cnt)}
\textbf{Question:} Which room has the most doors?\\
\textbf{Options:}\par
\casechoices{\textbf{A.} Video 1}{\textbf{B.} Video 2}{\textbf{C.} Video 3}{\textbf{D.} Video 4}
\textbf{Ground truth:} B

\caseturn{Turn 1 / GlobalAgent}{\analyze{}}
\textbf{Reason:} To determine which room has the most doors, I need to count the doors visible in each video. [...]\par
\textbf{video\_id:} \texttt{video\_1, video\_2, video\_3, video\_4}\par
\textbf{instruction:} Count the total number of doors visible in the room shown in this video. [...]\par
\textbf{Report / video\_1:} There is one visible door in the room. It is a wooden, single-leaf interior door located on the right wall. [...]\par
\textbf{LocalAgent / video\_2:} local steps\par
\textbf{Step 1 / action:} \observe{}\par
\textbf{where:} \texttt{[[0.0, 54.48]]}\par
\textbf{fps:} \texttt{1.5}\par
\textbf{what:} Count the total number of doors visible in the room. [...]\par
\textbf{Observation:} There are two doors visible in the room. 1. A white wooden door is located on the left wall, near the washing machine and radiator (visible at 0:04, 0:23, 0:45). 2. A [...] white wooden door is located on the back wall, leading to another room or hallway (visible at 0:00, 0:07, 0:29). Both doors appear to be standard hinged interior doors.\par
\textbf{Action:} \finish{}\par
\textbf{Report / video\_2:} There are two doors visible in the room. [...]\par
\textbf{Report / video\_3:} There is one door visible in the room. It is a wooden, single-panel interior door with a silver handle, located on the right wall. [...]\par
\textbf{Report / video\_4:} There is one visible door in the room. It is a wooden door located on the left side of the frame, partially open, leading into the bathroom. [...]\par
\caseturn{Turn 2 / GlobalAgent}{\answer{}}
\textbf{Reason:} [...] Video 2 has the highest count (2 doors). [...]\par
\nopagebreak[4]
\casefinal{B}

\end{casebox}

\subsubsection{Behavioral interpretation}
\begin{casebox}{agentteal}{Task: CrossVid (BU)}
\textbf{Question:} Which videos show behaviors directly caused by predator--prey interactions? Select all that apply.\\
\textbf{Options:}\par
\casechoices{\textbf{A.} Video 1}{\textbf{B.} Video 2}{\textbf{C.} Video 3}{\textbf{D.} Video 4}
\textbf{Ground truth:} C and D

\caseturn{Turn 1 / GlobalAgent}{\analyze{}}
\textbf{Reason:} I need to examine each video individually to determine if it shows behaviors directly caused by predator-prey interactions. [...]\par
\textbf{video\_id:} \texttt{video\_1, video\_2, video\_3, video\_4}\par
\textbf{instruction:} Watch the video and describe any interactions between animals. Specifically, identify if there is a predator chasing, attacking, or capturing prey, or if prey is fleeing from a predator. [...]\par
\textbf{Report / video\_1:} The video shows a group of sharks swimming together in deep blue water in a loose formation. No other animal species are clearly visible. [...]\par
\textbf{Report / video\_2:} The video shows a large barracuda swimming calmly near a coral reef. No other animals are visible in the clip, and there is no evidence of predator-prey interactions [...] capturing, or fleeing. [...]\par
\textbf{LocalAgent / video\_3:} local steps\par
\textbf{Step 1 / action:} \observe{}\par
\textbf{where:} \texttt{[[0.0, 1.62]]}\par
\textbf{fps:} \texttt{10.5}\par
\textbf{what:} Are there any animals in the video, and if so, what interactions (such as chasing, attacking, capturing, or fleeing) are occurring between them?\par
\textbf{Observation:} Yes, there are two animals visible in the video: a snake and a fish. The interaction is a predation event where the snake is capturing and consuming the fish. [...]\par
\textbf{Action:} \finish{}\par
\textbf{Report / video\_3:} The video shows a clear predation event between two animals: a snake and a fish. [...]\par
\textbf{Report / video\_4:} The video shows a clear predator-prey interaction between a snake and a fish. [...]\par
\caseturn{Turn 2 / GlobalAgent}{\answer{}}
\textbf{Reason:} [...] Therefore, videos 3 and 4 show the requested behaviors.\par
\nopagebreak[4]
\casefinal{CD}

\end{casebox}

\subsubsection{Temporal and procedural reasoning}
\begin{casebox}{agentteal}{Task: CrossVid (PSS)}
\textbf{Question:} Given five shuffled segments of a cooking video, output their correct order.\\
\textbf{Ground truth:} \texttt{2->4->5->3->1}

\caseturn{Turn 1 / GlobalAgent}{\analyze{}}
\textbf{Reason:} To determine the correct order of the shuffled cooking video segments, I first need to understand the content of each individual segment. [...]\par
\textbf{video\_id:} \texttt{segment\_1, segment\_2, segment\_3, segment\_4, segment\_5}\par
\textbf{instruction:} Describe the visual content of this cooking video segment, including any ingredients, actions, or cooking steps shown.\par
\textbf{LocalAgent / segment\_1:} local steps\par
\textbf{Step 1 / action:} \observe{}\par
\textbf{where:} \texttt{[[0.0, 11.01]]}\par
\textbf{fps:} \texttt{1.5}\par
\textbf{what:} What ingredients, actions, and cooking steps are shown in this cooking video segment?\par
\textbf{Observation:} The video shows a chef in a white uniform plating sushi rolls. [...]\par
\textbf{Action:} \finish{}\par
\textbf{Report / segment\_1:} The video shows a chef in a white uniform plating sushi rolls. [...]\par
\textbf{Report / segment\_2:} The video shows a chef in a professional kitchen preparing sushi. Visible ingredients include a sheet of nori (seaweed) with sesame seeds, a bed of white rice, [...]\par
\textbf{Report / segment\_3:} The video shows a professional kitchen scene where a chef in a white coat is preparing food on a white cutting board while a woman in a dark sleeveless top observes. [...] it is compact. [...]\par
\textbf{Report / segment\_4:} The video shows a chef assembling a sushi roll on a white countertop. The base is a sheet of nori (seaweed) with sesame seeds, topped with white rice. [...]\par
\textbf{Report / segment\_5:} The video shows a chef preparing sushi rolls in a professional kitchen. Visible ingredients include white rice, sliced avocado, cucumber strips, and orange-colored fish [...] assembling, topping, wrapping, and shaping the sushi roll. [...]\par
\caseturn{Turn 2 / GlobalAgent}{\watch{}}
\textbf{Reason:} To determine the correct chronological order of the cooking segments, I need to compare the visual progression of the sushi-making process [...]\par
\textbf{videos:} \texttt{[\{"video\_id": "segment\_1", "clip": [0.0, 11.01]\}, \{"video\_id": "segment\_2", "clip": [0.0, 18.99]\}, \{"video\_id": "segment\_3", "clip": [0.0, 17.02]\}, \{"video\_id": "segment\_4", "clip": [0.0, 5.01]\}]}\par
\textbf{instruction:} Analyze the visual continuity of the sushi preparation process across these segments to determine the correct chronological order. Specifically, look for the progression of steps: placing fillings on nori, rolling the sushi, adding toppings, shaping with a mat/wrap, and final plating. [...]\par
\textbf{Observation / observer(VLM):} [...] Here is a logical sequence based on the typical steps involved in making sushi: [...] Therefore, the correct chronological order of the video segments is: **Segment 1, Segment 2, Segment 3, Segment 4** The answer is: **1, 2, 3, 4**\par
\caseturn{Turn 3 / GlobalAgent}{\watch{}}
\textbf{Reason:} The previous joint-watch analysis (Event 2) incorrectly identified the content of the segments based on a generic template [...]\par
\textbf{videos:} \texttt{[\{"video\_id": "segment\_2", "clip": [0.0, 18.99]\}, \{"video\_id": "segment\_4", "clip": [0.0, 5.01]\}, \{"video\_id": "segment\_5", "clip": [0.0, 41.98]\}, \{"video\_id": "segment\_3", "clip": [0.0, 17.02]\}]}\par
\textbf{instruction:} Analyze the visual continuity of the sushi preparation to determine the correct order. Specifically, compare the state of the sushi roll in segment\_2 (filling placement), segment\_4 (rolling), segment\_5 (adding toppings and shaping with mat), segment\_3 (shaping with plastic wrap), and segment\_1 (plating). Identify which segment logically [...]\par
\textbf{Observation / observer(VLM):} [...] **Segment 2:** This segment shows the initial assembly. The chef places the fillings (a reddish mixture and a green vegetable) onto a sheet of rice and nori. This is the first step in creating the roll. [...] Therefore, the correct chronological order that follows the sushi-making process from start to finish is: **Segment 2, Segment 4, Segment 5, Segment 3, Segment 1**\par
\caseturn{Turn 4 / GlobalAgent}{\answer{}}
\textbf{Reason:} The joint-watch analysis in Event 3 provides a clear and logical chronological order for the sushi preparation process based on visual continuity: Segment 2 (assembly/filling placement) -\textgreater{} Segment 4 (rolling) -\textgreater{} Segment 5 (adding outer toppings and shaping) -\textgreater{} Segment 3 (final shaping with plastic wrap) -\textgreater{} Segment 1 (plating). [...]\par
\nopagebreak[4]
\casefinal{2->4->5->3->1}

\end{casebox}

\begin{casebox}{agentteal}{Task: MVU-Eval (TR)}
\textbf{Question:} What is the correct order of these ten videos if arranged by the time when the actions occur?\\
\textbf{Options (video IDs):}\par
\begin{caseoptions}
\item[\textbf{A.}] \texttt{4, 10, 2, 7, 6, 3, 5, 8, 9, 1}
\item[\textbf{B.}] \texttt{4, 6, 7, 8, 2, 3, 10, 1, 5, 9}
\item[\textbf{C.}] \texttt{9, 6, 8, 1, 5, 2, 10, 7, 4, 3}
\item[\textbf{D.}] \texttt{1, 6, 3, 4, 2, 7, 10, 8, 5, 9}
\end{caseoptions}
\textbf{Ground truth:} C

\caseturn{Turn 1 / GlobalAgent}{\analyze{}}
\textbf{Reason:} To determine the correct chronological order of the 10 videos, I first need to understand the content and actions in each video. [...]\par
\textbf{video\_id:} \texttt{video\_1, video\_2, video\_3, video\_4, video\_5, video\_6, video\_7, video\_8, video\_9, video\_10}\par
\textbf{instruction:} Describe the main action or event occurring in this video, including any visible time indicators, sequence markers, or context that suggests its position in a larger timeline.\par
\textbf{LocalAgent / video\_1:} local steps\par
\textbf{Step 1 / action:} \observe{}\par
\textbf{where:} \texttt{[[0.0, 6.01]]}\par
\textbf{fps:} \texttt{1.5}\par
\textbf{what:} What is the main action or event occurring in the video, including any visible time indicators, sequence markers, or context that suggests its position in a larger timeline?\par
\textbf{Observation:} A person pours a mixture of chopped red onion, green chilies, garlic, and liquid from a glass measuring cup into a clear zip-top plastic bag [...] The action occurs continuously from 0.0 to 6.01 seconds in the source video.\par
\textbf{Action:} \finish{}\par
\textbf{Report / video\_1:} The main event in the video is a person preparing a marinade or seasoning mix. [...]\par
\textbf{Report / video\_2:} The main action in the video is a person vigorously massaging and mixing raw meat (reddish-brown), yellow and green bell pepper pieces, and a liquid marinade [...]\par
\textbf{Report / video\_3:} The video presents a static close-up of three soft-shell tacos on an orange plate, filled with guacamole, diced vegetables, shredded cheese, and a white sauce. [...]\par
\textbf{Report / video\_4:} The video depicts a person slicing cooked meat on a white cutting board with a pink edge. [...]\par
\textbf{Report / video\_5:} The main action in the video is a person sealing a clear plastic zip-top bag containing marinated ingredients [...]\par
\textbf{Report / video\_6:} The video depicts the preparation of raw meat for marination. At the start (0.0s), several pieces of red meat are shown in a white tray. [...]\par
\textbf{Report / video\_7:} The video depicts the process of searing a large, seasoned raw steak in a hot cast iron skillet on a stovetop. [...]\par
\textbf{Report / video\_8:} The video depicts a person preparing a marinade for raw meat, likely tuna, inside a clear zip-top plastic bag. [...] The action is continuous and focused on adding citrus elements to the bag with the meat, suggesting this is an early stage in a larger food preparation process.\par
\textbf{Report / video\_9:} The main action in the video is a hand handling, lifting, turning, and spreading thin slices of raw red meat (likely beef) arranged on a white styrofoam tray [...]\par
\textbf{Report / video\_10:} The main action in the video is a hand sealing a clear plastic zip-top bag containing marinated meat, lime slices, and other ingredients [...]\par
\caseturn{Turn 2 / GlobalAgent}{\watch{}}
\textbf{Reason:} [...] I will watch key clips to confirm the ingredient addition order and the sealing steps.\par
\textbf{videos:} \texttt{[\{"video\_id": "video\_8", "clip": [0.0, 23.02]\}, \{"video\_id": "video\_1", "clip": [0.0, 6.01]\}, \{"video\_id": "video\_5", "clip": [0.0, 21.02]\}, \{"video\_id": "video\_10", "clip": [0.0, 5.0]\}]}\par
\textbf{instruction:} Analyze the sequence of actions in these clips to determine the chronological order of the food preparation process. Specifically, identify the order of: 1) adding ingredients to the bag (citrus vs. onion mixture), 2) sealing the bag, and 3) massaging the bag. [...]\par
\textbf{Observation / observer(VLM):} [...] In **video\_10**, the bag is fully sealed and contains all the components (meat, citrus, and onion mixture). The person is actively massaging and squeezing the bag to distribute the marinade evenly over the meat. This is the final action in this sequence. [...] Therefore, the correct chronological order of the food preparation process as depicted in the clips is: **video\_8 $\rightarrow$ video\_1 $\rightarrow$ video\_5 $\rightarrow$ video\_10**\par
\caseturn{Turn 3 / GlobalAgent}{\watch{}}
\textbf{Reason:} To determine the full chronological order of all 10 videos, I need to establish the sequence of the remaining videos (2, 3, 4, 6, 7, 9) [...]\par
\textbf{videos:} \texttt{[\{"video\_id": "video\_9", "clip": [0.0, 12.01]\}, \{"video\_id": "video\_6", "clip": [0.0, 22.02]\}, \{"video\_id": "video\_7", "clip": [0.0, 68.0]\}, \{"video\_id": "video\_2", "clip": [0.0, 21.02]\}]}\par
\textbf{instruction:} Analyze the state of the food and actions in these clips to determine their chronological order in the cooking process. Specifically, compare: 1) The state of the meat (raw on tray vs. in bag vs. cooking vs. cooked/sliced). 2) The contents of the bag (empty, adding citrus, adding liquid, sealed, massaging). 3) The final presentation. [...]\par
\textbf{Observation / observer(VLM):} [...] Therefore, the complete chronological order of all the video clips is: **video\_2, video\_8, video\_1, video\_5, video\_10, video\_7, video\_9, video\_4, video\_6**\par
\caseturn{Turn 4 / GlobalAgent}{\answer{}}
\textbf{Reason:} The previous joint-watch analysis (Event 3) provided a chronological order: video\_2, video\_8, video\_1, video\_5, video\_10, video\_7, video\_9, video\_4, video\_6. However, this analysis missed video\_3 (the final plated tacos) and contained some inconsistencies in the description of the marinating steps compared to Event 2. [...] Option C has 5, 2, 10. This is plausible if 2 and 10 are both massaging steps. Then 7 (cook), 4 (slice), 3 (serve). This looks correct. [...]\par
\nopagebreak[4]
\casefinal{C}

\end{casebox}

\subsubsection{Contextual event inference}
\begin{casebox}{agentteal}{Task: CrossVid (PI)}
\textbf{Question:} Given the beginning and ending of a movie clip, what is most likely to happen in the middle?\\
\textbf{Options (abridged):}\par
\begin{caseoptions}
\item[\textbf{A.}] The wounded man threatens the shopkeeper.
\item[\textbf{B.}] The middle-aged man bribes a security guard.
\item[\textbf{C.}] The wounded man discovers GPS data in the shop.
\item[\textbf{D.}] The shopkeeper warns the middle-aged man to flee.
\item[\textbf{E.}] The wounded man follows the middle-aged man's vehicle to a forest cabin.
\item[\textbf{F.}] The security guard supplies a forged document.
\end{caseoptions}
\textbf{Ground truth:} E

\caseturn{Turn 1 / GlobalAgent}{\analyze{}}
\textbf{Reason:} To determine the most likely middle event, I first need to understand the context, characters, and actions in the beginning and ending clips [...]\par
\textbf{video\_id:} \texttt{beginning}\par
\textbf{instruction:} Describe the scene, characters, and key actions in this video clip. Note any injuries, locations, interactions between characters, and the overall situation at the start.\par
\textbf{video\_id:} \texttt{ending}\par
\textbf{instruction:} Describe the scene, characters, and key actions in this video clip. Note the final location, the condition of the characters, any discoveries made, and the outcome of the events.\par
\textbf{LocalAgent / beginning:} local steps\par
\textbf{Step 1 / action:} \observe{}\par
\textbf{where:} \texttt{[[0.0, 128.0]]}\par
\textbf{fps:} \texttt{1.5}\par
\textbf{what:} Describe the scene, characters, and key actions in this video clip. [...]\par
\textbf{Observation:} The video opens with a woman driving a car with a young boy in the passenger seat; both appear distressed. [...] The overall situation suggests urgency and distress, with the injured man seeking something or someone at the pawn shop before departing.\par
\textbf{Action:} \finish{}\par
\textbf{Report / beginning:} [...] The overall situation conveys urgency and distress, with the injured man appearing to seek something or someone at the pawn shop before departing.\par
\textbf{Report / ending:} The video depicts a scene in a dimly lit, institutional setting with cinder block walls and fluorescent lighting. [...] The video concludes with the two men still seated at the table, continuing their discussion in the same institutional room.\par
\caseturn{Turn 2 / GlobalAgent}{\answer{}}
\textbf{Reason:} The beginning video shows a wounded man visiting a pawn shop and then observing a white pickup truck, suggesting he is tracking someone. The ending video shows a prisoner (likely the man from the truck or related) being interrogated in a holding cell. The missing middle segment likely connects the tracking action to the capture. [...] I will select E as the most likely event.\par
\nopagebreak[4]
\casefinal{E}

\end{casebox}

\subsection{Failure cases and analysis}
\label{app:failure-cases}

\subsubsection{Multi-view comparison}
\begin{casebox}{agentamber}{Task: MVU-Eval (Cmp)}
\textbf{Question:} Which car is traveling the fastest among the left front, front, and right front views?\\
\textbf{Options:}\par
\casechoices{\textbf{A.} Red Car}{\textbf{B.} Blue Car}{\textbf{C.} Brown Car}{\textbf{D.} White Car}
\textbf{Ground truth:} C\par
\textbf{MVAgent prediction:} \texttt{D}\par

\caseturn{Turn 1 / GlobalAgent}{\analyze{}}
\textbf{Reason:} [...] Since I don't know which video ID corresponds to which view, I will analyze all videos to identify the view type and the cars present in each.\par
\textbf{video\_id:} \texttt{video\_1, video\_2, video\_3, video\_4, video\_5, video\_6}\par
\textbf{instruction:} Identify the camera view (e.g., left front, front, right front) and list all cars visible, including their colors and any observable motion cues like speed or direction.\par
\textbf{LocalAgent / video\_1:} local steps\par
\textbf{Step 1 / action:} \observe{}\par
\textbf{where:} \texttt{[[0.0, 4.0]]}\par
\textbf{fps:} \texttt{5.5}\par
\textbf{what:} Identify the camera view (e.g., left front, front, right front) and list all visible cars, including their colors and any observable motion cues such as speed or direction.\par
\textbf{Observation:} Camera view: right front (passenger-side perspective looking forward and slightly right). Visible cars: 1) White box truck moving forward in the adjacent lane, [...] No other moving vehicles are observed in the clip.\par
\textbf{Action:} \finish{}\par
\textbf{Report / video\_1:} Camera view: right front (passenger-side perspective looking forward and slightly right). Visible cars: 1) White box truck moving forward in the adjacent lane, [...]\par
\textbf{Report / video\_2:} The camera view is front-facing from inside a vehicle, looking through a rain-streaked windshield. [...]\par
\textbf{Report / video\_3:} Camera view: right front (passenger-side perspective looking forward and slightly right). Visible cars and motion cues: 1) Black sedan in the right lane moving forward [...]\par
\textbf{Report / video\_4:} Camera view: right front (passenger-side perspective looking forward and slightly right). Visible cars: 1) White van with 'fios' logo on side, moving forward [...]\par
\textbf{Report / video\_5:} Camera view: right front (passenger-side perspective looking forward and slightly right). Visible cars and motion cues: 1) Silver pickup truck [...]\par
\textbf{Report / video\_6:} The camera view is front-facing, looking down a wet city street. [...] All vehicles appear stationary or moving slowly due to wet road conditions.\par
\caseturn{Turn 2 / GlobalAgent}{\watch{}}
\textbf{Reason:} [...] I will select video\_2 (front), video\_6 (front), and video\_1 (right front) to compare the motion of vehicles in these views.\par
\textbf{videos:} \texttt{[\{"video\_id": "video\_2", "clip": [0.0, 4.0]\}, \{"video\_id": "video\_6", "clip": [0.0, 4.0]\}, \{"video\_id": "video\_1", "clip": [0.0, 4.0]\}]}\par
\textbf{instruction:} Compare the motion speeds of the moving vehicles in the selected videos. [...] Determine which specific car (by color) is traveling the fastest among the vehicles visible in the 'left front', 'front', and 'right front' views. [...]\par
\textbf{Observation / observer(VLM):} [...] - **Video 2:** - **Camera View:** Front [...] - **Video 6:** - **Camera View:** Left Front [...] The motion speeds in all three videos are moderate, with no specific car standing out as the fastest. [...]\par
\caseturn{Turn 3 / GlobalAgent}{\answer{}}
\textbf{Reason:} The joint-watch result (Event 2) explicitly states that among the vehicles visible in the 'left front' (Video 6), 'front' (Video 2), and 'right front' (Video 1) views, none of the cars (Red, Blue, Brown, or White) are traveling at a high speed, and no specific car stands out as the fastest. [...] Therefore, by elimination, the White Car is the only candidate for 'fastest' among the options, as it is the only one moving. [...]\par
\nopagebreak[4]
\casefinal{D}

\end{casebox}

\begin{failureanalysis}
The selected V2, V6, and V1 correspond to back, front, and back-right views in the dataset paths, omitting the required front-left and front-right sources. The final answer also treats a moving white vehicle as the fastest, although the joint observation establishes no relative-speed ordering.
\end{failureanalysis}

\subsubsection{Functional alignment and temporal localization}
\begin{casebox}{agentamber}{Task: CrossVid (FSA)}
\textbf{Question:} Which step in Video 2 is functionally equivalent to the step shown between 213 and 222 seconds in Video 1? Answer with a time interval.\\
\textbf{Ground truth:} 34--42 s\par
\textbf{MVAgent prediction:} \texttt{66,68}\par

\caseturn{Turn 1 / GlobalAgent}{\analyze{}}
\textbf{Reason:} [...] I will analyze this clip to describe the cooking action performed.\par
\textbf{video\_id:} \texttt{video\_1}\par
\textbf{instruction:} Describe the specific cooking action or step shown between 213 seconds and 222 seconds. [...]\par
\textbf{LocalAgent / video\_1:} local steps\par
\textbf{Step 1 / action:} \observe{}\par
\textbf{where:} \texttt{[[213.0, 222.0]]}\par
\textbf{fps:} \texttt{1.5}\par
\textbf{what:} What specific cooking action or step is shown between 213 and 222 seconds? [...]\par
\textbf{Observation:} [...] He is pouring a clear liquid from a large plastic bottle into a stainless steel pan that is on the burner. [...]\par
\textbf{Action:} \finish{}\par
\textbf{Report / video\_1:} Between 213 and 222 seconds, a man is shown pouring a clear liquid from a large plastic bottle into a stainless steel pan on a stovetop. The pan contains a simmering mixture of chopped vegetables and possibly meat or sauce. [...]\par
\caseturn{Turn 2 / GlobalAgent}{\analyze{}}
\textbf{Reason:} I have identified the action in Video 1 (213-222s) as pouring a clear liquid into a simmering pan. [...] I will request a single-video analysis of Video 2 to describe its cooking steps, specifically looking for liquid addition.\par
\textbf{video\_id:} \texttt{video\_2}\par
\textbf{instruction:} Describe the cooking steps shown in this video. Specifically, identify any step where a liquid is poured or added to a pan or pot containing other ingredients. [...]\par
\textbf{LocalAgent / video\_2:} local steps\par
\textbf{Step 1 / action:} \observe{}\par
\textbf{where:} \texttt{[[0.0, 126.36]]}\par
\textbf{fps:} \texttt{1.5}\par
\textbf{what:} Identify all cooking steps where a liquid is poured or added to a pan or pot containing other ingredients. [...]\par
\textbf{Observation:} At 00:41, the chef pours heavy cream from a white bowl into a hot pan containing melted butter. At 01:06, the chef pours vodka from a clear glass into the pan, [...]\par
\textbf{Action:} \finish{}\par
\textbf{Report / video\_2:} [...] First, at 00:41, heavy cream is poured from a white bowl into a hot pan that contains melted butter. Second, at 01:06, vodka is poured [...]\par
\caseturn{Turn 3 / GlobalAgent}{\watch{}}
\textbf{Reason:} The step in Video 1 (213s-222s) involves pouring a clear liquid from a large plastic bottle into a simmering pan. In Video 2, the summary identifies [...] to ensure no other step in Video 2 better matches the visual characteristics (e.g., bottle shape, liquid clarity, steam) of the Video 1 segment.\par
\textbf{videos:} \texttt{[\{"video\_id": "video\_1", "clip": [213.0, 222.0]\}, \{"video\_id": "video\_2", "clip": [40.0, 70.0]\}]}\par
\textbf{instruction:} Compare the pouring action in video\_1 (213s-222s) with the liquid pouring steps in video\_2 (specifically around 00:41, 01:06, and 01:08). Identify which step in video\_2 is functionally equivalent to the step in video\_1, focusing on the type of liquid (clear vs opaque), the container used, and the state of the food in the pan.\par
\textbf{Observation / observer(VLM):} In video\_1, the chef is pouring a clear liquid from a bottle into a pan that already contains some food. [...] - **At 00:41:** The chef pours a white, opaque liquid (likely cream or milk) from a small white bowl into a pan. [...] Therefore, there is no step in video\_2 that matches the pouring action in video\_1 in terms of the type of liquid, the container used, and the state of the food in the pan.\par
\caseturn{Turn 4 / GlobalAgent}{\analyze{}}
\textbf{Reason:} The joint-watch analysis concluded that no step in video\_2 is functionally equivalent to the step in video\_1 (213s-222s) [...]\par
\textbf{video\_id:} \texttt{video\_2}\par
\textbf{instruction:} Re-examine the step around 01:06 in video\_2. Specifically, describe the liquid being poured: is it clear or opaque? [...]\par
\textbf{video\_id:} \texttt{video\_1}\par
\textbf{instruction:} Re-examine the step between 213s and 222s in video\_1. [...]\par
\textbf{LocalAgent / video\_2:} local steps\par
\textbf{Step 1 / action:} \observe{}\par
\textbf{where:} \texttt{[[1.0, 2.0], [104.0, 110.0]]}\par
\textbf{fps:} \texttt{1.5}\par
\textbf{what:} At the step around 01:06, describe the liquid being poured: is it clear or opaque? [...]\par
\textbf{Observation:} ,\par
\textbf{uncertainty:} The requested event (pouring liquid at 01:06) is not visible in the supplied clip, which only shows a static logo for 'RAO'S CAESARS PALACE' from source time 1.0--2.0 seconds. [...]\par
\textbf{Observation:} , the video clip provided (source interval 104.0--110.0 seconds) does not show any liquid being poured. [...]\par
\textbf{uncertainty:} The requested event (pouring liquid around 01:06) is not present in the supplied clip (which corresponds to source time 104--110 seconds). The clip only shows plating, not cooking or liquid addition.\par
\textbf{Step 2 / action:} \observe{}\par
\textbf{where:} \texttt{[[1.0, 2.0], [104.0, 110.0], [105.0, 111.0]]}\par
\textbf{fps:} \texttt{1.5}\par
\textbf{what:} At the step around 01:06, describe the liquid being poured: is it clear or opaque? [...]\par
\textbf{error:} \observe{} ranges must be chronological and non-overlapping.\par
\textbf{Step 3 / action:} \observe{}\par
\textbf{where:} \texttt{[[1.0, 2.0], [104.0, 110.0], [105.0, 111.0], [60.0, 70.0]]}\par
\textbf{fps:} \texttt{1.5}\par
\textbf{what:} At the step around 01:06, describe the liquid being poured: is it clear or opaque? [...]\par
\textbf{error:} \observe{} ranges must be chronological and non-overlapping.\par
\textbf{Step 4 / action:} \observe{}\par
\textbf{where:} \texttt{[[60.0, 70.0]]}\par
\textbf{fps:} \texttt{2.5}\par
\textbf{what:} At the step around 01:06, describe the liquid being poured: is it clear or opaque? [...]\par
\textbf{Observation:} At source time 01:06 (clip time \textasciitilde{}6.0s), a thick, opaque red liquid (likely tomato sauce) is being poured from a metal bowl into a black frying pan. [...]\par
\textbf{Action:} \finish{}\par
\textbf{Report / video\_2:} At approximately 01:06, a thick, opaque red liquid (likely tomato sauce) is poured from a metal bowl into a black frying pan. [...] The detailed visual evidence from the targeted range confirms the red opaque liquid from a metal bowl.\par
\textbf{Report / video\_1:} Between 213s and 222s, a clear liquid is poured from a large plastic bottle into a stainless steel pan on the stovetop. [...]\par
\caseturn{Turn 5 / GlobalAgent}{\answer{}}
\textbf{Reason:} [...] Video 2's broad scan identified a clear liquid (vodka) poured from a glass at 01:06, which is functionally equivalent to deglazing or adding liquid to a sauce. Although a detailed re-examination at 01:06 noted red liquid, the broad scan explicitly listed the vodka step at 01:06 and tomato sauce at 01:08, suggesting the vodka pour occurs immediately before or at the start of that interval. Given the requirement for a clear liquid equivalent, the vodka step at 01:06 is the only candidate.\par
\nopagebreak[4]
\casefinal{66,68}

\end{casebox}

\begin{failureanalysis}
Subsequent queries follow the initial clear-liquid candidate outside the annotated interval. The range [40,70] covers only the interval's end, while [60,70] misses it. Reinspection revises the event description without reopening the temporal search. A source-frame check shows a pour from a transparent cup at 36 s within the annotation, leaving the reported ``heavy cream at 41 s'' unverified.
\end{failureanalysis}

\subsubsection{Fine-grained spatial relations}
\begin{casebox}{agentamber}{Task: MVU-Eval (Cmp)}
\textbf{Question:} What specific edit operation transforms Video 1 into Video 2?\\
\textbf{Options (abridged):} Add a snake that is\par
\begin{caseoptions}
\item[\textbf{A.}] coiling around debris near the man;
\item[\textbf{B.}] perched on a nearby structure;
\item[\textbf{C.}] slithering behind the man;
\item[\textbf{D.}] resting on the background fence;
\item[\textbf{E.}] coiled on the ground away from the man;
\item[\textbf{F.}] hiding under the man's boot;
\item[\textbf{G.}] lying flat on the ground far from the man;
\item[\textbf{H.}] crawling away from the man;
\item[\textbf{I.}] coiling on the ground near the man;
\item[\textbf{J.}] moving toward the man.
\end{caseoptions}
\textbf{Ground truth:} I\par
\textbf{MVAgent prediction:} \texttt{A}\par

\caseturn{Turn 1 / GlobalAgent}{\watch{}}
\textbf{Reason:} To determine the specific edit operation, I must visually compare Video 1 and Video 2 to identify the added element (a snake) and its precise location and behavior relative to the man and surroundings. [...]\par
\textbf{videos:} \texttt{[\{"video\_id": "video\_1", "clip": [0.0, 5.0]\}, \{"video\_id": "video\_2", "clip": [0.0, 5.1]\}]}\par
\textbf{instruction:} [...] Describe exactly what object has been added in Video 2 that is not present in Video 1, including its location relative to the man, its posture (e.g., coiling, slithering, resting), and any interaction with nearby objects like debris, structures, or the fence.\par
\textbf{Observation / observer(VLM):} In Video 2, a snake has been added to the scene. The snake is located on the ground to the right of the man's boots. It is in a coiled posture [...] The snake is positioned near some debris on the ground but does not seem to be interacting directly with any structures or the fence in the background.\par
\caseturn{Turn 2 / GlobalAgent}{\answer{}}
\textbf{Reason:} [...] This matches option A: 'Add a snake coiling around a piece of debris near the man.' The description of the snake being 'near some debris' and 'coiled' aligns with the option text.\par
\nopagebreak[4]
\casefinal{A}

\end{casebox}

\begin{failureanalysis}
GlobalAgent interprets the coiled snake near debris as coiling around it and selects A. Option I matches the reported ground-level location near the man; the error lies in mapping the joint observation to an option.
\end{failureanalysis}

These cases do not estimate success rates or isolate component effects. Model-allocation and component comparisons use their reported configurations; incomplete raw remote ablation records limit causal interpretation of small score differences.

\end{document}